\documentclass{article}

\usepackage[preprint]{neurips_2022}

\usepackage{natbib}

\usepackage{microtype}
\usepackage{booktabs} 
\usepackage{lipsum}
\usepackage{subfig} 
\usepackage{xcolor}

\usepackage{hyperref}

\usepackage{wrapfig}

\usepackage{amsmath}
\usepackage{amssymb}
\usepackage{mathtools}
\usepackage{amsthm}
\usepackage{listings}
\usepackage[capitalize,noabbrev]{cleveref}
\crefname{section}{Sec.}{Secs.}
\Crefname{section}{Section}{Sections}
\Crefname{table}{Table}{Tables}
\crefname{table}{Tab.}{Tabs.}

\usepackage[inline]{enumitem}

\newcommand{\x}{\mathbf{x}}
\newcommand{\tx}{\tilde{\mathbf{x}}}
\renewcommand{\d}[1]{\ensuremath{\operatorname{d}\!{#1}}}
\newcommand{\bfw}{\mathbf{w}}
\newcommand{\bftheta}{{\boldsymbol{\theta}}}
\newtheorem{remark}{Remark}
\newtheorem{obsn}{Observation}
\newtheorem{lemma}{Lemma}
\DeclareMathOperator*{\argmin}{arg\,min}
\newcommand{\norm}[1]{\left\lVert#1\right\rVert}
\newcommand{\expct}{\mathbb{E}}
\newcommand{\regn}{\mathcal{R}}

\usepackage[textsize=tiny]{todonotes}

\title{Using Intermediate Forward Iterates for Intermediate Generator Optimization}

\author{%
Harsh Mishra$^{1}$ \quad Jurijs Nazarovs$^{2}$ \quad Manmohan Dogra$^1$ \quad Sathya N. Ravi$^1$ \\
$^1$Department of Computer Science, University of Illinois at Chicago\\
 $^2$Department of Statistics, University of Wisconsin Madison\\
}

\begin{document}

\maketitle



\begin{abstract} Score-based models have recently been introduced as a richer framework to model distributions in high dimensions and are generally more suitable for generative tasks. In score-based models, a generative task is formulated using a parametric model (such as a neural network) to directly learn the gradient of such high dimensional distributions, instead of the density functions themselves, as is done traditionally. From the mathematical point of view, such gradient information can be utilized in reverse by stochastic sampling to generate diverse samples. However, from a computational perspective, existing score-based models can be efficiently trained only if the forward or the corruption process can be computed in closed form. By using the relationship between the process and layers in a feed-forward network, we derive a backpropagation-based procedure which we call {\em Intermediate Generator Optimization} to utilize intermediate iterates of the process with negligible computational overhead. The main advantage of IGO is that it can be incorporated into any standard autoencoder pipeline for the generative task. We analyze the sample complexity properties of IGO to solve downstream tasks like Generative PCA. We show applications of the IGO on two dense predictive tasks viz., image extrapolation, and point cloud denoising. Our experiments indicate that obtaining an ensemble of generators for various time points is possible using first-order methods. 

\end{abstract}

\section{Introduction}

\label{sec:intro}
Modular priors defined using such generative models have enabled a broad spectrum of applications in our vision community. For instance, it is now a well-accepted notion that ill-posed-ness in many inverse problems such as image denoising, inpainting, and compressed sensing can be mitigated effectively in practice using such priors. Generative Adversarial Networks (GANs) were first shown to be successful in generating high-resolution realistic natural images \cite{wu2019gp}, and biomedical images \cite{harutyunyan2001high} for augmentation purposes. Variational Autoencoders (VAE) is a popular alternative that is based on minimizing the {\em distortion} given by integral metrics such as KL divergence.  In both GANs and VAEs, the learning problem coincides -- we seek to learn the process of generating new samples based on latent space modeled as random distributions \cite{higgins2016beta,razavi2019generating,zhang2019d}. Applications enabled by DGMs in vision include style transferring \cite{zhu2017unpaired}, inpainting \cite{demir2018patch}, image restoration and manipulation \cite{pan2021exploiting}. {VAE architectures have been successfully deployed in temporal prediction settings owing to their robustness properties \cite{nazarovs2021variational,rubanova2019latent}.

Score-based models are popular when it comes to deep generative models, and in many cases have outperformed
the previous state-of-the-art, GANs \cite{dhariwal2021diffusion, song2021scorebased, ho2021cascaded}. They have also been extensively used for Text-to-Image generation, \cite{https://doi.org/10.48550/arxiv.2112.10741, https://doi.org/10.48550/arxiv.2204.06125}. These models transform the given data into a known prior by slowly perturbing it using a Stochastic Differential Equation (SDE) and then undo the perturbation process by solving the SDE in reverse-time. This reverse-time SDE is approximated by a time-dependent neural network. For efficiently training these neural networks, the prior distribution is chosen to be a Gaussian distribution, mean and variance as a function of $t$ can be written in closed-form expressions that can be evaluated efficiently.

\begin{figure*}[!t]
    \centering
    {\includegraphics[width=\textwidth]{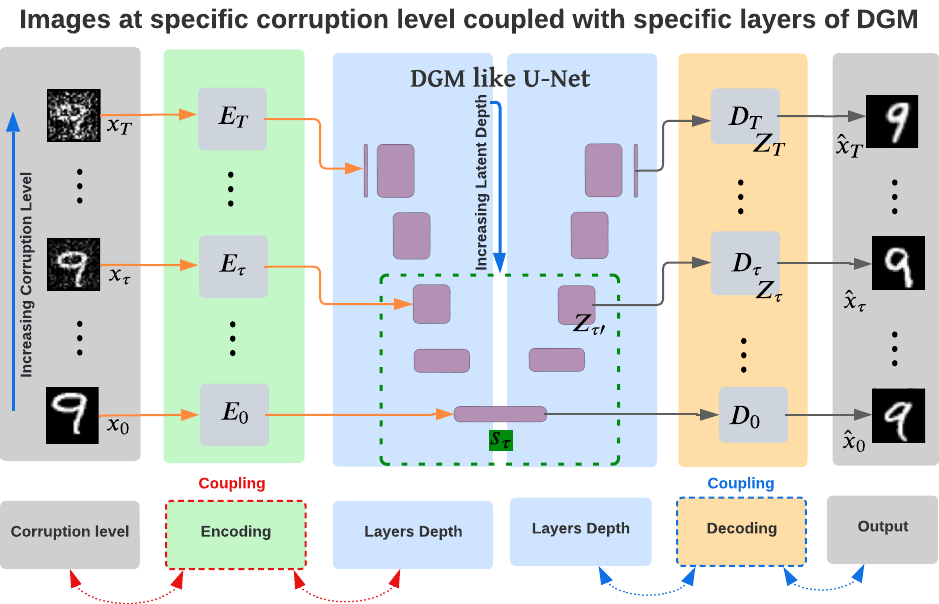}}
    \caption{ Figure shows our idea of coupling iterates from a given corruption process with Generators. }%
    \vspace{-15pt}
    \label{fig:flow_arch}%
\end{figure*}

{\bf How to train Score Based Models?}  In score based models, the goal is to learn the {\em gradients} $\nabla_{\x} \log p_t(\x)$ by training a 
time-dependent score based model $s_\bftheta(\x, t)$. In particular, we seek to solve the following distributional optimization problem \cite{song2021scorebased}:
\begin{align}
\small
\bftheta^* = \argmin_\bftheta 
   \expct_t\left[\lambda(t) \expct_{\x(0)}\expct_{\x(t) \mid \x(0) } L(s_{\theta},x(t),t)
   \right],%
    \label{eqn:training}
\end{align}
where $L(s_{\theta},x(t),t)$ is usually chosen as the squared loss given by $ \left[\norm{s_\bftheta(\x(t), t) - \nabla_{\x(t)}\log p_{t}(\x(t) \mid \x(0))}_2^2 \right]$.  We refer to the loss function in \cref{eqn:training} as the standard (or usual) loss function. To understand the loss function consider the standard Kernel Density Estimation (KDE) procedure for the moment. KDE is a smooth estimator of Probability density function, and is known to converge faster than histograms (function values), in terms of sample complexity, see  \cite{wasserman2010statistics}. While parameters of KDEs are usually tuned to maximize the likelihood of observed data, using higher order (or finer) information such as gradients have been studied recently in small scale settings, see \cite{sasaki2017estimating,kim2019uniform}. Note that if $x(t)\in\mathbb{R}^n$, then $\nabla_{x(t)} \log p_{t}(x(t) \mid x(0))\in\mathbb{R}^n$, whereas $\log p_{t}(x(t) \mid x(0))\in\mathbb{R}$, so training $s_{\theta}$ to fit gradients $\nabla_{x(t)}\log p_{t}(x(t) \mid x(0))$ may be slightly  more challenging than simply maximizing the likelihood $\log p_{t}(x(t) \mid x(0))$.

In Score based DGMs, the generator $G$ learns the {\em gradient} of the log-likelihood function explicitly instead of the density function itself. There are two benefits associated with this approach: (i) statistical range of the generator can be naturally improved during sampling since gradients are often more informative (smooth function values can be approximated using gradients by Taylor's theorem, and (ii) computationally tractable alternative to kernel density estimation since it does not require us to estimate the normalizing constant which is usually intractable in high dimensions.

}

{\paragraph{Problem Statement.}
In this work we propose ways to mitigate two technical challenges, (i) training Score-based Generative models under non-Gaussian corruption procedures, while (ii)  increasing the range of such Generative $G$ models using intermediate iterates of smooth SDEs while training $s_{\theta}$. We show how our novel technique Intermediate Generator Optimization, can be used to increase the diversity of $G$ and solve various inverse problems, when given the access to the forward process of SDEs where the forward operator cannot be computed easily.
}



{\paragraph{Our Contributions.} Motivated by our applications in various large scale vision applications, (i) we propose a new fully differentiable framework that explicitly models the corruption process within DGMs, and which can be trained using standard backpropagation procedures like SGD and its variants; (ii) we analyze the statistical properties of our proposed framework called Intermediate Generator Optimization (IGO) using recently introduced Intermediate Layer Optimization (ILO) framework \cite{daras2021intermediate} for solving inverse problems and moreover; (iii)  we show the utility of our procedure with extensive results on several image generation and prediction tasks. In the challenging, three dimensional point cloud setting, we identify potentially beneficial regularizers for improving the robustness profile of denoisers. We show how to formulate and implement recently introduced projected power method (PPower) \cite{liu2022generative} within our IGO framework (code - \url{https://github.com/harshm16/intermediate_generator_optimization} ).} 

{{\paragraph{Roadmap.} In Section \ref{sec:disc}, we show how efficient discretization strategies can be used for non-Gaussian corruption processes.
In Section \ref{sec:dgmdp}, using Information Bottleneck Principle, we introduce the notion of an intermediate generator regularizer for utilizing iterates of a discrete process during training. We argue that our regularizer can be efficiently optimized and analyze its sample complexity for linear inverse problems. We discuss various vision use cases of IGO such as image generation, dense extrapolation, and point cloud denoisers along with our experimental results in Section \ref{sec:exps}.}}








\section{Non-Gaussian corruption via Discretization}\label{sec:disc}
\paragraph{Basic Setup.} A Stochastic Differential Equations (SDE), where $f$ and $g$ are the drift and the diffusion functions respectively, and $w$ is the Standard Brownian Motion is represented by \ref{eqn:forward_sde}:\begin{align}
    \d\x = f(\x,t)\d t+g(t)\d w. \label{eqn:forward_sde}
\end{align}
We can map data to a noise distribution using the forward SDE in \cref{eqn:forward_sde} and reverse the SDE for generation using: \begin{align}
\d \x = [f(\x,t)-g(t)^2\nabla_{\x} \log p_t(\x)] \d t + g(t) d \bar{\bfw}, \label{eqn:backward_sde}
\end{align}
where, $\bar{\bfw}$ is the Brownian motion when time flows backward from $T$ to $0$, and $\d t$ is an infinitesimal negative timestep. After training using \cref{eqn:training}, we may simply replace the log-likelihood in the reverse SDE by our model $s$ in \cref{eqn:backward_sde} for sampling purposes. In essence, an approximate model $s$ will allow us to generate diverse, yet realistic samples using a small amount of independent random noise component $g$. We present a discussion about why the SDE-based loss function automatically induces diversity in samples generated by a generative model in \cref{sec:discussion_sde}.

\paragraph{Our Assumption on Forward Process.}
For the SDE model given in \cref{eqn:forward_sde}, typically the drift function $f$ is chosen to be affine, thus bypassing the need to obtain $p_{t}(x(t) \mid x(0))$. We consider  learning the trajectory information under cases when  $f$ is not necessarily affine viz.,  Lotka-Volterra and Arnold's Cat Map as $f$. To that end, we relax the  assumption that we have access to an efficient discretization of the process in  \cref{eqn:forward_sde}. Specifically, we say that the forward SDE \cref{eqn:forward_sde} can be efficiently computed using the Euler-Maruyama (EM) method. We will use $\tx$ to denote the approximation to $\x$ provided by the EM discretization. For a time-varying multiplicative process defined by $\d\x_t = a(\x_t,t) \d t + b(\x_t,t) \d w_t$, where $a$ and $b$ are smooth functions similar to $f$ and $g$ in \cref{eqn:forward_sde}, but $a$ is not affine, the EM algorithm executes the following iterations: \begin{align}
\tx_{j+1} = \tx_j + a(\tx_j,t)\Delta t + b(\tx_j,t)z\sqrt{\Delta t}, \label{eqn:EM}
\end{align}
where $\tx_j$ denotes the $j-$th iterate and $z\sim\mathcal{N}(0,1)$ is a sample drawn uniformly at random from the standard normal distribution. Note that the process defined in \cref{eqn:forward_sde} satisfies our assumption here since we allow $b$ to depend on $x_t$. The discretization provided by \cref{eqn:EM} enables us to simulate the SDE and subsequently sample from $p_{t}(x(t) \mid x(0))$.

\section{Efficient DGMs for Discrete Processes}\label{sec:dgmdp}
To provide guarantees on DGM-based downstream tasks such as solving inverse problems, we propose a new loss function to train the score-based model $s$ based on individual trajectories of samples corrupted by the discretized process defined in \cref{eqn:EM}. We begin by writing the empirical finite sample form of the optimization problem in \cref{eqn:training} as follows, \begin{align}
    \min_{\bftheta} \sum_{\tau=t_1}^{t_T} \lambda\left(\tau\right)\sum_{i=1}^T \expct_{\x_i(\tau)|\x_i} L\left(s_{\theta},\x_i\left(\tau\right),t\right),\label{eq:erm}
\end{align}
where $T$ represents the discretization size, and $0\leq t_i\leq 1~\forall~i=1,\dots,T$. To solve the optimization problem in \cref{eq:erm}, the most popular choice is to use first-order backpropagation type methods. Importantly, the worst-case complexity of solving \cref{eq:erm} scales linearly with the discretization size $T$, which is intractable in large-scale settings. By the chain rule, the efficiency of such algorithms strongly depends on the ease of evaluating the derivative of the forward operator, denoted by the conditional expectation. While previous works assume that there is a closed-form solution to evaluate the conditional expectation, such an assumption is invalid in our examples discussed above. If $T=1$, then we may simply simulate the dataset $\{\x_i\}$ using \cref{eqn:EM} to obtain $\tx_i(t_1)$, and use it to compute the loss, and backpropagate.

\subsection{Handling $T=2$ case using IBP} Now we consider the  setting in which we have access to only one intermediate iterate $\tx_{\tau}$ where $\tau<t$ for some arbitrary $t\sim\mathcal{U}(0,1)$ in our trajectory. Naively, we can train two DGMs in parallel, one each for $\tx_t$ and $\tx_{\tau}$, thus incurring twice the memory and time complexity, including resources spent for hyperparameter tuning. We propose a simpler alternative through the lens of the so-called Information Bottleneck Principle (IBP) -- the de-facto design principle used in constructing standard Autoencoders architectures such as U-Net. 

\paragraph{Coupling Forward Processes with Intermediate Iterates.} In feedforward learning, the main result due to IBP is that layers in a neural network try to {\em compress} the input while maximally preserving the relevant {\em information} regarding the task at hand \cite{goldfeld2020information}. For designing neural networks, this corresponds to choosing a sequence of transformations $T_l$ such that the distance between successive transformations $d(T_l,T_{l-1})$ is not too big. In practice, we can ensure this by simply choosing dimensions of layerwise weight matrices by a decreasing function of layers (or depth). Intuitively, the idea is that if dimensions of $T_l$ is a constant factor of $T_{l-1}$, then at optimality (after training), when random input passes via $T_l$, a constant factor of noise available when it passed through $T_{l-1}$ will be removed, in expectation. 



\paragraph{IGO for $\tx_{\tau}$ iterates.} By using the correspondence between the forward process defined in \cref{eqn:EM} and intermediate layers in a DGM, we will now define our loss function for intermediate iterates $\tx_{\tau}$. For simplicity, we drop the subscript $\textbf{i}$.
We define the regularization function $\regn$ as follows, \begin{align}
    \regn(\tx_{\tau},t)= L\left(D_{\tau}\circ s_{\tau} \circ E_{\tau},\tx_{\tau},t\right),\label{eqn:loss}
\end{align}
where $s_{\tau}$ (see dark green box in Figure \ref{fig:flow_arch}) denotes the {\em restriction} of the score based model $s$ to iterate $\tx_{\tau}$, $E_{\tau}$ denotes a shallow encoder for $\tx_{\tau}$, and similarly for the decoder $D_{\tau}$. 
We will refer to the DGM defined by $D_{\tau}\circ s_{\tau}\circ E_{\tau}$ as the {\bf Intermediate Generator}, see Figure \ref{fig:flow_arch}. Intuitively, the regularization function $\regn$ is defined to modify the parameters of the model $s$ to a specific set of connected layers given by $s_{\tau}$. To see this, first, note that the overall parameter space $\bftheta$ can be seen as a product space over layers $\theta_l$. Then, at any given training iteration, $\regn$ function {\em restricts} the update to layers that are suitable for decoding $\tx_{\tau}$.



\paragraph{Interpreting $\mathcal{R}$.}
{Using a small step size in EM guarantees us that a larger class of SDEs can be simulated 
{see \cite{doi:10.1137/18M1170017}) }, so $\tilde{x} \sim x$. So, let us fix a sufficiently small $\Delta t$, so $\tx\approx\x$ almost everywhere at any time $t$. Now, when the forward process \cref{eqn:EM} is a {\em Markov} process, that is, increments of $g$ are independent of $t$, then the intermediate iterates do not carry any more information. In fact, in practice, it may be detrimental to the training process if the gradient directions are not sufficiently correlated.  Unfortunately, this assumption is not true in our discrete setting considered here since both the drift $a$, and diffusion $b$ functions are time-varying. In this case, $\mathcal{R}$ simply tries to revert the time-dependent noise process in the relevant part of the network $s_{\tau}$ using appropriate $E_{\tau}$, and $D_{\tau}$. Our regularization function provides time-dependent noise information explicitly by using the intermediate iterate $\tx_{\tau}$ during training in appropriate parts of DGM. This makes sure that our IGO framework can also be extended to Energy based models, which are known to perform as well as Score based DGMs \cite{salimans2021should}, for image generation.}
\paragraph{Extension to Multiple Iterates $T>2$.} It is straightforward to extend our regularization function when we are given more than one intermediate iterate from the discretized EM algorithm by,\begin{align}
    R_{0}^T(\tx)=\sum_{\tau=t_1}^{t_T} L\left(D_{\tau}\circ s_{\tau} \circ E_{\tau},\tx_{\tau},t\right).\label{eq:multipleiterates}
\end{align}

\paragraph{Implementation.} IGO utilizes less computational resources in the following sense -- if $E_{\tau}$, and $D_{\tau}$, relatively shallow networks (as shown in \cref{fig:flow_arch}), then the cost of computing gradients with $\tx_{\tau}$ is negligible compared to $\tx$, thus achieving cost savings by design.

\begin{figure*}[!t]
    \centering
     \subfloat[\textbf{(L:R)} Generated by $Z_{T}$, $Z_{\tau}$ and $Z_{\tau\prime}$ respectively, \cref{sec:gpca}.]{\includegraphics[width=0.5\linewidth]{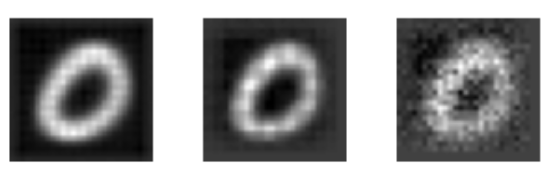} }
     \hfill
    \subfloat[Number of Principal Components required for Explaining Variance in \cref{sec:gpca}]{\includegraphics[width=0.4\linewidth]{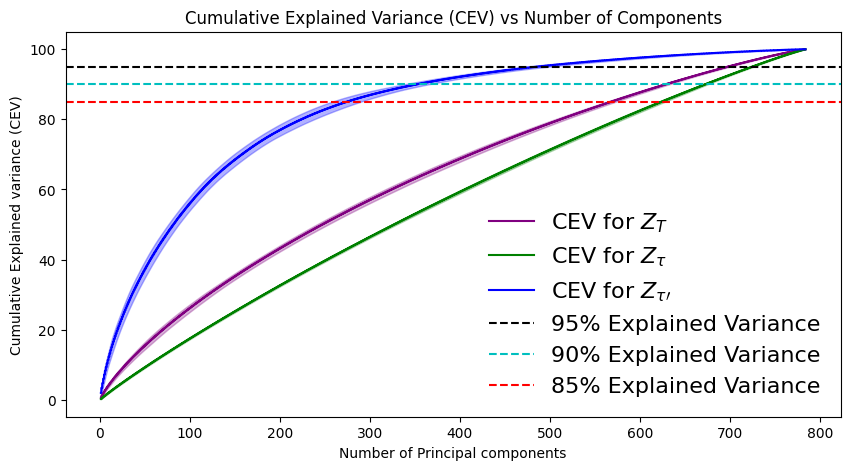} }
    
    \caption{Generative PCA with IGO, \cref{sec:gpca}
    }%
    \label{fig:generated_images_pca}%
    \vspace{-15pt}
    
\end{figure*}

\section{Experiments}\label{sec:exps}
In this section, we explore different applications of IGO across varying setups to showcase our idea of utilizing an intermediate iterate $\tx_{\tau}$. First, in \cref{sec:gpca} we show our IGO framework for Generative PCA, using a pre-trained network.
Next, we show the procedure to train IGO in a U-Net based architecture in \cref{sec:igo-unet}.
Then, show the utility of intermediate features for prediction. In \cref{sec:panel-ode}, we tackle the challenging trajectory prediction tasks, in particular, we modify the ODE to take inputs from the intermediate, and final iterates during training. Finally, In \cref{sec:pcd-exps} we demonstrate the use of IGO in a {\em dense} {prediction} task of denoising posed as a supervised learning problem. Here, a multi-layer perceptron is utilized as a score-based model for denoising purposes.

\subsection {Generative PCA with IGO} 
\label{sec:gpca}

\subsubsection{Setup Details} 
Here we follow the Projected Power (PPower) method proposed by \cite{liu2022generative}, to solve the eigenvalue problem using pre-trained generators capable of generating MNIST digits. The basic idea of the classic power method for the eigenvalue problem, is to repeatedly apply the eigenvector matrix to a starting vector (noise) and then normalize it to converge to the eigenvector associated with the largest eigenvalue. The key difference between the PPower method compared to the classic power method is that it uses an additional projection operation to ensure that the output of each iteration lies in the range of a generative model. We perform experiments by optimizing the noise vector in 3 different cases. Firstly, we optimize the noise vector $Z_{T}$, used for generation at the last decoding layer of the usual generator. Then, we optimize $Z_{\tau}$, for generation at the last decoding layer of the Intermediate Generator. Finally, we use $Z_{\tau\prime}$ to generate at a smaller dimension, as indicated in \cref{fig:flow_arch}. The projection step for each case is solved by using the generated samples from the respective decoders, using an Adam optimizer with 100 steps and a learning rate of 0.001.

\subsubsection {Experimental results}

Firstly, we see from \cref{fig:generated_images_pca} - (a), that our model was able to generate images in all the 3 cases.
The image generated by reconstructing $Z_{\tau\prime}$, the rightmost in \cref{fig:generated_images_pca} - (a), looks noisier, since the model's features at that stage were smaller in comparison to the other cases. We also analyze the Cumulative Explained Variance as a function of the number of components, in \cref{fig:generated_images_pca} - (b). We see that the images generated using a smaller dimension of noise vector $Z_{\tau\prime}$ use less principal components to explain variance as compared to $Z_{T}$ and $Z_{\tau}$.

\paragraph{Takeaway.}
By Theorem 3 in \cite{liu2022generative} we know that the reconstruction error is proportional to root dimensions of $Z$. Our experiments indicate that the smaller dimensions of Z in $Z_{\tau}$ and $Z_{\tau\prime}$ can also be used to generate images. Thus giving us lower reconstruction error as well as parameter savings of approximately $Z_{T}$ - $Z_{\tau\prime}$ dimensions.

\begin{figure}[!t]
    \includegraphics[width=1\columnwidth]{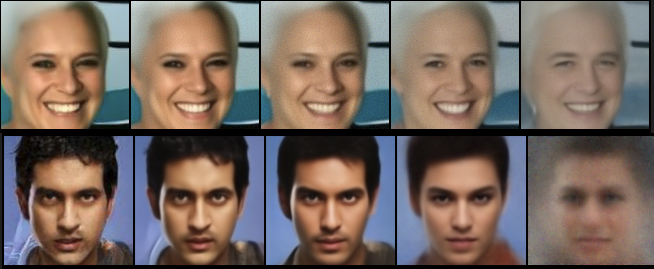}
    \caption{
    Generated Samples for \cref{sec:igo-unet}
    }
    \label{fig:example}
     \vspace{-15pt}
\end{figure}

\subsection {IGO for training DGMs with non-Gaussian Drift} 
\label{sec:igo-unet}
\subsubsection{Setup Details} 
In a standard score-based generation, the transition kernel is assumed to be a multivariate Gaussian distribution. Given a clean image $\x(0)$, we can apply the kernel using closed form evaluation to obtain the noisy image $\x(t)$ for random $t\sim\mathcal{U}(0,1)$. Neural networks are then used to estimate the score, also known as the time-dependent gradient field, to reverse the corruption process. 

Here, we start off with explicitly modeling our corruption process using a non-affine SDE with Arnold's cat map as the drift, to attain an intermediate iterate $\tx_{\tau}$, by setting up Eqn. \eqref{eqn:EM}. Our goal here is to utilize our proposed Intermediate Generator Optimization, for training purposes. To do so, we simulate the forward SDE till some random time $t$ and store an (additional) intermediate iterate for every trajectory to obtain $\x_{t}$ and $\tx_{\tau}$ ({$\tau = t/2$}). The two iterates are then used to set the training objective based on Eqn. \eqref{eq:erm} with our proposed regularizer in Eqn. \eqref{eqn:loss}. As per \cref{fig:flow_arch}, the intermediate iterate $\tx_{\tau}$ has its own intermediate pathway $s_{\tau}$ into our overall model $s$.

\subsubsection {Experimental results}
We choose the CelebA dataset to test the generative capabilities of our IGO formulation. Similar to \cite{song2021scorebased} we also use a U-net \cite{ronneberger2015unet} architecture as the backbone of our model. The parameters of which are shared across time using sinusoidal position embeddings \cite{vaswani2017attention}. During the forward pass, our simulated iterates $\x_{t}$ and $\tx_{\tau}$ are passed into the network, using their specific pathways, refer \cref{fig:flow_arch}. For the iterate $\tx_{\tau}$ we use the intermediate encoder $E_{\tau}$ and an intermediate decoder $D_{\tau}$. The iterates $\x_{t}$ and $\tx_{\tau}$ are jointly used to train the network using a convex combination of loss. We use the scalar $1 - \alpha$ to refer the weight given to the gradient of the regularization function $\mathcal{R}$, \cref{eqn:loss}. So, the lesser the value of $\alpha$, the more the weight of the intermediate iterates in the objective function for training the score network. The trained score model is then provided to the RK45 ODE Solver (\emph{scipy}) for sampling. The top row of \cref{fig:example} shows the progressive samples generated using the intermediate layers, while the bottom row shows the samples from the usual layers. The rightmost images in \cref{fig:example} are the closest to noise. The model used to generate images in \cref{fig:example} was trained using $\alpha = 0.5$. We utilize the pytorch implementation of Fretchet Inception Distance (FID) \cite{NIPS2017_8a1d6947} score provided by \cite{Seitzer2020FID}. The images from the intermediate layers have an average FID of $3.5$, while the images from the usual layers have an FID of $3.3$, similar to \cite{song2021scorebased}. All the comparisons were made using 192 feature dimension and using the model with number of residual blocks per resolution as 4. We provide results on the effect of $\alpha$, using results on MNIST data in \cref{sec:mnistsupp}.

\paragraph{Takeaway.} Our results show the advantage of using the IGO framework, as the intermediate layers are able to generate images of almost similar quality (FID) as the usual layers, while providing a shorter pathway for lesser corrupted iterates of a noise process.

\subsection{IGO on Panel Data}
\label{sec:panel-ode}
\subsubsection{Setup Details}
\begin{figure*}[!t]

    \centering
        \subfloat[Incorporation of intermediate iterates in Mixed Effect Neural ODE for analyzing the dynamics of Panel data.]{\includegraphics[width=0.47\textwidth]{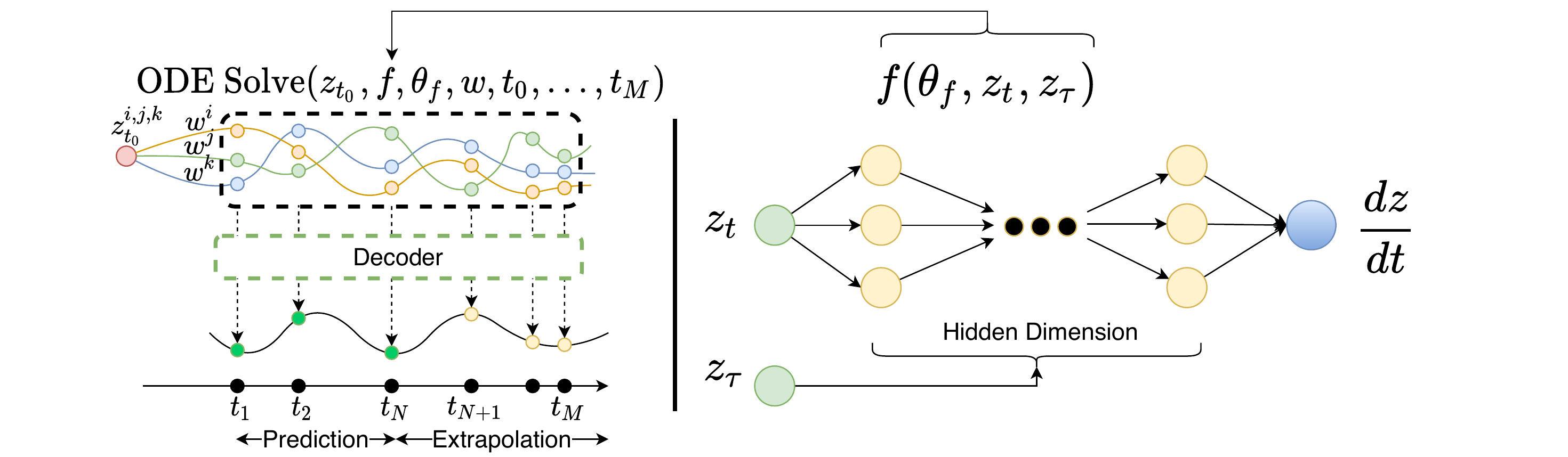} }
        \hfill
        \hfill
    \subfloat[One of the Hopper samples: \textbf{Top} is the true data, \textbf{Bottom} is the prediction. Before the black line is observed data, after it are the extrapolated samples previously not seen by the model.]{\includegraphics[width=0.47\textwidth]{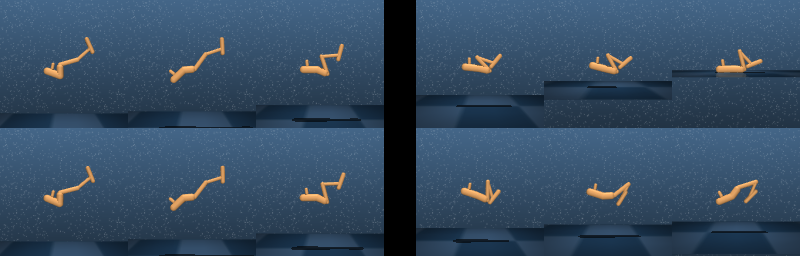} }
    
    \caption{IGO on Panel Data, \cref{sec:panel-ode}}%
    \vspace{-15pt}
    \label{fig:paneldata}%
    
\end{figure*}

One of the applications of our proposed Intermediate Generator Optimization is modeling the trajectory of a Differential Equation, which describes the dynamic process. We use the setup in \cite{nazarovs2021variational}, that is, we consider that dynamic process is modeled by changes in the latent space $z$ as $\dot{z}(t) = D(z, t)\mathbf{w}$, where $D(z, t)$ is a neural network and $\mathbf{w}$ is a corresponding mixed effect (random projection describing flexibility (stochasticity) of dynamics). We propose to extend $D(z, t)$ with our IGO, as $D(z_t, z_{\tau})$ using a neural network $f$, see \cref{fig:paneldata} - (a) for details on loss computation with $z_{\tau}.$

\subsubsection {Experimental results}
We evaluate our approach on two temporal vision datasets, representing the dynamics of Panel Data. Namely, we apply our proposed method to the variation of MuJoCo Hopper and Rotation MNIST.

\paragraph{MuJoCo Hopper}
The dataset represents the process with simple Newtonian physics, which is defined by the initial position, velocity, and the number of steps.
To generate data we randomly chose an initial position and sample velocity vector uniformly from $[-2,2]$. We evaluate our model on interpolation (3 steps) and extrapolation (3 steps) and provide visualization of one of the experiments in \cref{fig:paneldata} - (b). The MSE for interpolation was 0.0289, while for Extrapolation was 0.2801. In comparison ODE2VAE \cite{nazarovs2021variational}'s interpolation MSE is at 0.0648.

\paragraph{Rotation MNIST}
We evaluate our approach on a more complicated version of the rotating MNIST dataset. We construct a dataset by rotating the images of different handwritten digits and reconstructing the trajectory of the rotation in interpolation and extrapolation setups. For a sampled digit we randomly choose an angle from the range $[-\pi/4, \pi/4]$ and apply it at all time steps. In addition, to make our evaluation more robust and closer to a practical scenario, we spread out the initial points of the digit, by randomly rotating a digit by angles from $-\pi/2$ to $\pi/2$. 
The generated 10K samples of different rotating digits for $20$ time steps were split into two equal sets: interpolation and extrapolation. Like the previous experiment, MuJoCo Hopper, we evaluate our model on interpolation (10 steps) and extrapolation (10 steps). In this case, the MSE for interpolation and extrapolation were 0.0082 and 0.1545 respectively. In comparison NODE \cite{nazarovs2021variational} is 0.0074 and 0.1661 respectively. Whereas MEODE \cite{nazarovs2021variational} is 0.0057 and 0.1641 respectively. Refer to \cref{sec:panelsupp} for implementation details.

\paragraph{Takeaway.}
 Clearly, we can see that by introducing IGO in the model, we get a good generation ability even for extrapolation in the future time steps.

\begin{figure*}[!t]
    \centering
     \subfloat[IGO Model for 3D point cloud denoising.]{\includegraphics[width=0.5\linewidth]{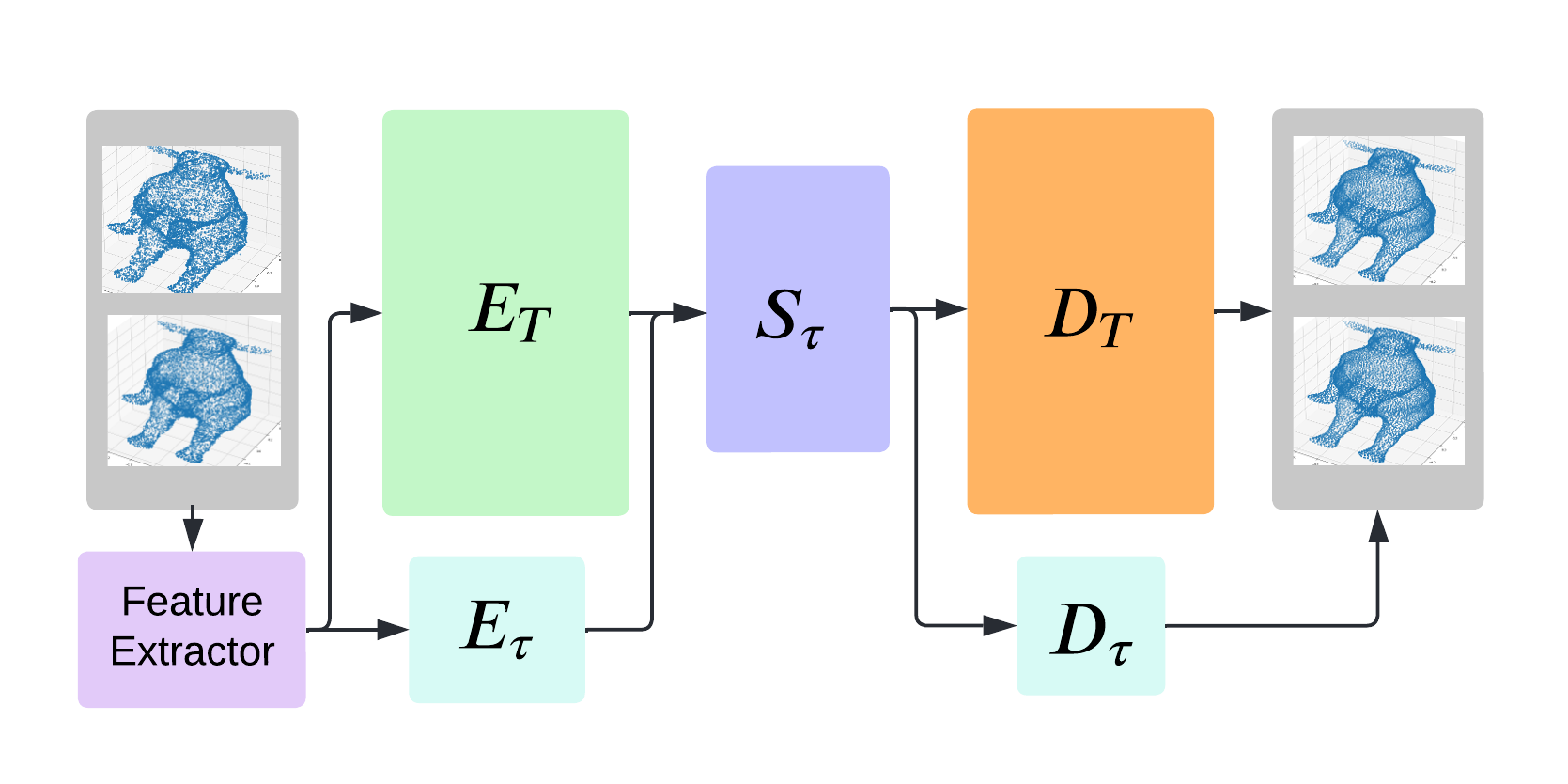} }
     \hfill
    \subfloat[Cosine Similarity]{\includegraphics[width=0.23\linewidth]{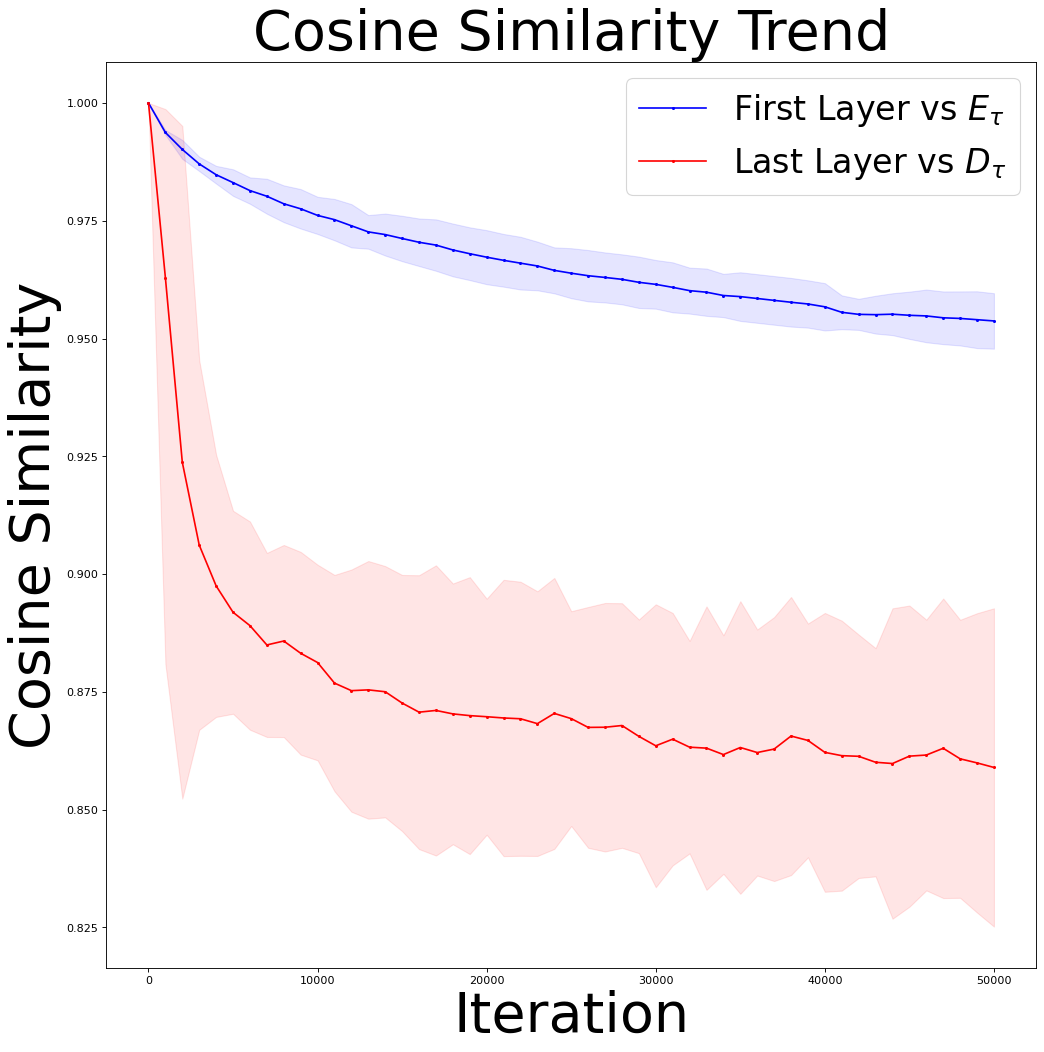} }
    \hfill
    \subfloat[Pairwise Euclidean distance]{\includegraphics[width=0.23\linewidth]{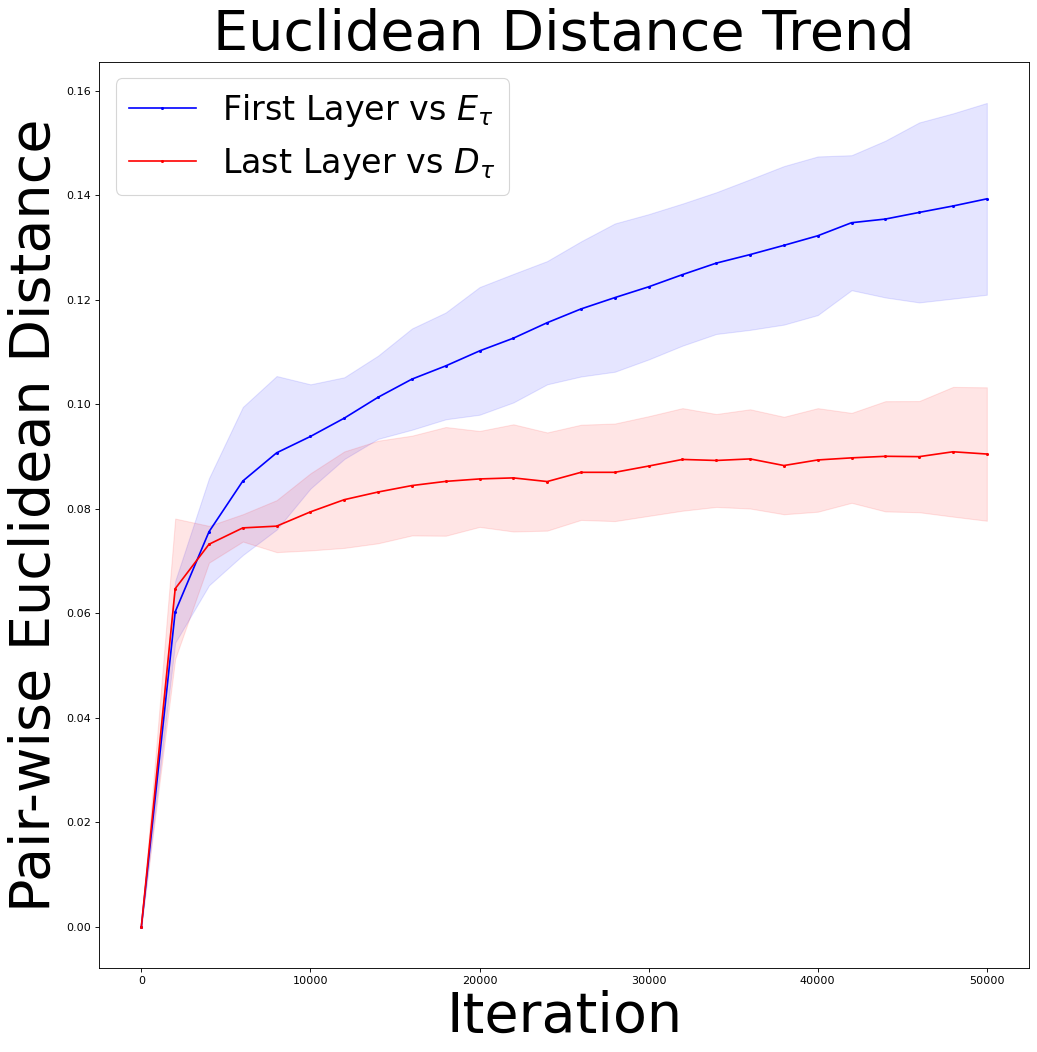} }%
    \caption{IGO in 3D Processing. Figures \textbf{(b)} and \textbf{(c)} compare the weights of $E_{T}$ against $E_{\tau}$, and $D_{T}$ against $D_{\tau}$. }%
    \label{fig:3dpointcloud}%
    \vspace{-15pt}
    
\end{figure*} 

\subsection{IGO in 3D Processing} 

\label{sec:pcd-exps}
\subsubsection{Setup Details} We use a standard deep learning-based point cloud denoiser that comprises a Feature Extraction Unit and a Score Estimation Unit. The feature extractor is tasked to learn local as well as non-local features for each point in the noisy point cloud data, provided as the input. Here, the score estimator provides point-wise scores, using which the gradient of the log-probability function can be computed. We adopt the same feature extraction unit as well as the Score Estimation Unit used in \cite{Luo_2021_ICCV}. Here we introduce intermediate iterates to test whether we can generate different versions of the denoised samples while using a pretrained score-based model provided by \cite{Luo_2021_ICCV}, see Figure \ref{fig:3dpointcloud} - a).

\subsubsection {Experimental results} 
At the beginning of the denoising step, we have with us the given noisy input, $\x_{t}$ and its intermediate iterate $\tx_{\tau}$. Our architecture is modelled using a 3 layer MLP as in \cite{Luo_2021_ICCV}, and the newly added intermediate layers, $E_{\tau}$ and $D_{\tau}$. The newly added layers replicate the already present first and the last layer perceptrons but are only dedicated to the intermediate layers. We initialize the weights of $E_{\tau}$ and $D_{\tau}$ to be half of the weights of the pre-trained layers. The training was done using a convex combination of loss, defined using the intermediate and the final iterate, $\x_{t}$ and $\tx_{\tau}$. The architecture can be seen in \cref{fig:3dpointcloud} - (a). We use $\alpha$ to denote this convex combination hyperparameter. The lesser the value of $\alpha$, the more the weight of the intermediate iterates in the objective function for training the score network. The network was tested using the PU-Net test set provided by \cite{Luo_2021_ICCV}. We provide comparisons to different baselines using Chamfer Distance (CD) \cite{fan2016point} and Point-to-Mesh Distance (P2M) \cite{ravi2020accelerating} as the metric, in \cref{sec:igo-3dsupp}.

\subsubsection {Finding new Generators using $X_{\tau}$ } \label{sssec:cosine_sim}
Here we validate our hypothesis that we can train diverse generators using intermediate iterate. In order to quantify the spread of our overall model, we use the cosine similarity metric suggested for generalization purposes \cite{jin2020does}. Here, we compute the cosine similarities between the weights of $E_{T}$ and $E_{\tau}$ (also, $D_{T}$ and $D_{\tau}$). Recall, $D_{T}$ is the usual decoder which is the last layer, and $D_{\tau}$ is its corresponding intermediate decoder. We also calculate the pair-wise Euclidean distance between the respective weights, see Figure \ref{fig:3dpointcloud} - b), c).

\paragraph{Takeaway.} We observed that as we fine-tune our model the cosine similarity between $D_{T}$ and $D_{\tau}$ decreases by $10 \%$ and the Euclidean between $E_{T}$ and $E_{\tau}$ increases by $15 \%$, in $50k$ iterations \textemdash two dissimilar decoders that perform well on the training dataset.

\section{Theoretical Analysis for Downstream Tasks}
We utilize the key result in ILO to analyze our procedure for downstream tasks. To that end, we assume that the parameters of $s$, $E_{\tau}$ and $D_{\tau}$ are fixed (given by a pre-trained model), and show that IGO is suitable for solving inverse problems under low sample settings. 

\paragraph{Necessary Condition on DGM.} We follow the same observation model as in \cite{daras2021intermediate} given by $y=Ax+\mathfrak{n}$ where $\mathfrak{n}$ is a random variable representing noise, and $x$ is the unknown. It is well known that when the measurement matrix $A$ satisfies certain probabilistic requirements, then it is possible to solve for $x$ using a single (sub-)gradient descent scheme \cite{candes2006near,tropp2015introduction}. This is specified using the S-REC condition in DGM-based prior modeling using CSGM algorithm \cite{bora2017compressed}. An example that satisfies the probabilistic requirements is when the entries in $A$ are distributed according to Gaussian distribution, as mentioned in  \cite{daras2021intermediate}.


To analyze $\mathcal{R}$ for prior modeling purposes, we assume that after training the generator has a compositional structure given by $G_1\circ G_2$, which is followed in most standard architectures. The following observation is crucial in analyzing IGO for compressive sensing purposes.
\begin{obsn} \label{obsn:range} {\bf (Range expansion due to $E_{\tau}$.)}
Setting the intermediate generator (of $s$) to be $G_{\tau}$, the range of our pre-trained model $G_1$ is increased to,\begin{align}
   \underbar{G}:=G_{\tau}(B_2^k(r_1)) \oplus E_{\tau}(B_2^{k_{\tau}}(r_{\tau})),
\end{align}
where $\oplus$ denotes the Minkowski Set sum, and $k_{\tau}$ corresponds to the dimensions of intermediate encoder associated with $\tx_{\tau}$.
\end{obsn}
With the increased range in Observation \ref{obsn:range}, we show that using the overall generator trained with IGO as a prior can be sample efficient in the following lemma.  We make the same assumptions as in Theorem 1. in \cite{daras2021intermediate} on the entries of $A$, and smoothness of $G_2$.
\begin{lemma}

{Assume that the Lipschitz constant $\underbar{G}$ is $\underbar{L}$, } and we run gradient descent to solve for the intermediate vector (as in CSGM). Assume that the number of measurements $m$ in $y$ is at least \begin{align}
    \min(k\log(L_1L_2r_1/{\delta}),k_{\tau}\log(\underbar{L}L_2r_{\tau}/{\delta}))+\log p,
\end{align}
where $p$ is the input dimensions of $G_2$ or the intermediate vector. Then, for a fixed $p$, we are guaranteed to recover an approximate $x$ with high probability (exponential in $m$).\label{lem:inv}
\end{lemma}
Please see \cref{sec:proof} for the proof of the signal recovery lemma \ref{lem:inv}. Our proof uses the (now) standard $\epsilon-$net argument over the increased range space \cite{vershynin2018high}.  In essence, if the Lipschitz constant of the intermediate generator $\underbar{L}$ is lower than the Lipschitz constant of $G_1$, then we get an improved sample complexity bound for IGO. 
Intuitively, lemma \ref{lem:inv} states that the number of samples required during downstream processing depends on two factors: \begin{enumerate*}
    \item {\em latent dimension size} denoted by the $\log p$ term which remains the same with or without intermediate iterates, and 
    \item {\em smoothness} denoted by the Lipschitz constants $L_1,\underbar{L},L_2$.
\end{enumerate*}

While we have no control over the size $p$, if the intermediate generator is smoother, and performs well on training data $y$, then gradient descent succeeds in recovering the ``missing" entries with fewer samples. In the case when intermediate iterates are not useful, then our framework allows us to simply ignore the intermediate generator without losing performance. In other words, our framework preserves the hardness of recovery -- easy problems remain easy. 

\paragraph{Lemma 1 on PCA:} 
To apply Lemma 1 to the problem of PCA with complex generative priors, like in \cref{sec:gpca}, we solve \cref{eq:gPCA}.
\begin{align}
 \hat{\mathbf{v}} := \max_{ \mathbf{w} \in R^n } \mathbf{w}^T \mathbf{V} \mathbf{w} \quad \text{s.t.} \quad \mathbf{w}  \in \mathrm{Range}(G_{T}  \cup G_{\tau}).\label{eq:gPCA}
\end{align}

Here, by $G_{T}  \cup G_{\tau}$ we mean the linear combination of distributions of the generators $G_{T}$ and $G_{\tau}$.

\begin{remark}[Difference between proposed IGO and ILO.]
The input of the ILO algorithm is pre-trained DGM with the goal of turning the noise distribution for better image generation using gradient descent schemes. In IGO, we use the knowledge of the forwarding process to optimize the parameters of the DGM to increase its range {\bf during} training.
\end{remark}
\vspace{-5pt}
\paragraph{Leveraging a diverse $G$  for generic Downstream Tasks.} 
Consider a feature extractor (or an encoder) induced by an appropriate SDE-guided DGM, that can extract {\em relevant} features from a mixture of two distributions, for example, training, and adversarial.  In this setting, if we know the distribution from which a given feature has been sampled, then we can hope to predict the sample at an optimal error rate by using a neural network with sufficient layers. When the mixtures have a natural dependency (given by SDE \cref{eqn:forward_sde}), then we may simply use weight sharing instead of training two separate models. We touched upon this in \cref{sec:pcd-exps} and \cref{fig:3dpointcloud} - a). Thus, achieving memory as well as some time savings. To further clarify, assume that a DGM has total of $P$ parameters. Then, training $|T|$ DGMs each for a time step of a discretization size $|T|$ would require $O(P|T|)$ parameters. However in our approach, say we require $p$ extra parameters for $E_{\tau}$, $D_{\tau}$, then the total number of parameters in our IGO framework is $O(P+(p-1)|T|) = O(P+|T|)$, for small values of $p$ – as is the case in our experimental settings – this can be a huge reduction in the number of parameters $O(P|T|)$ in standard framework vs $O(P+|T|)$ in our IGO framework.

\section{Conclusions} 
From the theoretical perspective, our results indicate that it is indeed possible to be more sample efficient while solving inverse problems using our IGO construction by slightly modifying standard assumptions made on DGMs. We can implement our regularizer on any end-to-end differentiable DGM with minor code modifications as shown in two 2D images and one 3D point cloud setting. Our experiments in the case of Eigenvalue problems also show that our framework and the landscape of its parameters has interesting properties and can be exploited for efficient optimization purposes and parameter savings. We believe that abstract ideas from dynamical systems are very much relevant in decision-making in vision settings as more and more pre-trained models are deployed on form factor and/or edge devices to make on-the-fly decisions. Our results provide us a shred of affirmative evidence that ideas from dynamical systems will be of at most importance when each such decision has nontrivial consequences (say due to the presence of adversarial noise).

\bibliography{egbib}
\bibliographystyle{abbrvnat}

\appendix
\clearpage
\section*{Appendix}
We include several appendices with additional details, proof, and additional results our experiments. First, in \cref{sec:proof} we show the proof for \cref{lem:inv}. We provide a detailed literature review in \cref{sec:literature_review}, which is followed by a Discussion on Improving range of generators using Stochastic Differential Equations in \cref{sec:discussion_sde}. We then provide the implementation details for using our proposed non-Gaussian drift functions in \cref{sec:applications}. In \cref{sec:expssupp}, we show additional results along with pseudocode for the experiments in \cref{sec:exps}. We also provide results from an additional experiment on training DGMs with IGO on MNIST dataset in \cref{sec:mnistsupp}.

\section{Proof of signal recovery Lemma 1}
\label{sec:proof}

\begin{proof}
Our proof follows the strategy detailed in \cite{daras2021intermediate}. In particular, we will use the Metric entropy concentration inequality for $\ell_2$ balls, and Maurey's empirical method (the Sudokov Minorization inequality) for $\ell_1$ ball, which are more accurate than the standard $\epsilon-$net argument. We will now explain the details.

In order to make any sample complexity guarantee using pretrained generator, we have to assume that we have to access to an algorithm that can solve inverse problem using a pretrained generator. For this purpose, we will use CSGM method for solving inverse problems using the generators that we have trained (that includes that the intermediate iterate). Observe that the intermediate generator has its own noise source that can be optimized using the CSGM method for inverse problem solving purposes. Hence, after training, we obtain two separate generators i.e., two independent noise sources. Following the notations in the main paper, for any decomposition of the overall $G=G_1\circ G_2$, we can define the respective true optimum in the extended range of the intermediate generator $G_{\tau}:=s_\tau\circ D_{\tau}$ be given as,\begin{align}
	\bar{z}_{\tau}^p=\arg\min_{z^p\in G_{\tau}(B_{2}^k(r_1))\oplus B_1^p(r_2)}\|x-G_2(z_p)\|,\label{eq:z-tau-true}
\end{align}
and its corresponding measurements optimum in the extended range of $G_{\tau}$ be given by $\tilde{z}_{\tau}^p$.  We will drop the superscript $p$ to avoid notation clutter. Then our results follows by noting that this intermediate generator $G_{\tau}$ satisfies the S-REC condition since the corresponding intermediate encoder $E_{\tau}$  and decoder $D_{\tau} $ are lipschitz. That is, following inequality (78) in \cite{daras2021intermediate} we have that,\begin{align}
	\|G_2(\bar{z}_{\tau})-G_2(\tilde{z})\|\leq \frac{4\| G_2(\bar{z})-x\|+\delta_{\tau}}{\gamma},
\end{align} where $\delta_{\tau}$ is a constant that depends (at most polynomially) on latent dimensions and $p$ only. We can now take the minimum of the sample complexity of both the generators since they both have independent noise sources. Finally, we have the desired result as claimed due to the S-REC property for nested $\ell_1$ ball of the intermediate generator. 
\end{proof}

\begin{figure*}[!t]
\centering
\includegraphics[width=\textwidth, height=10cm]{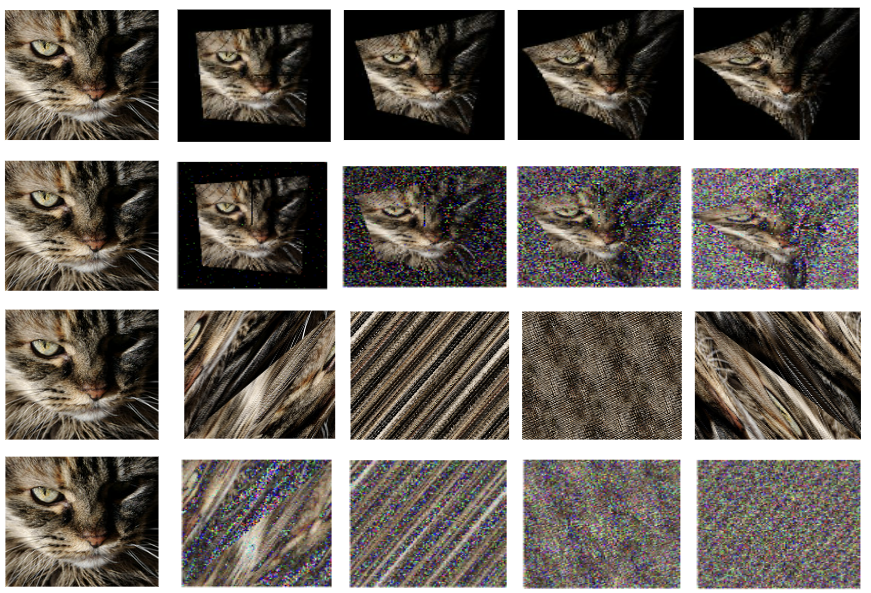}
\caption{ Data augmentation schemes. {\bf Row 1}: Lotka-Volterra. {\bf Row 2}: Lotka-Volterra + Gaussian Noise. {\bf Row 3}: Arnold's Cat Map. {\bf Row 4}: Arnold's Cat Map + Gaussian Noise.
\label{fig:augment}}
\end{figure*}

\section{Literature Review}
\label{sec:literature_review}

In recent years, there is a growth in literature, which focuses on utilizing ODE and SDE theory in Neural Networks\cite{chen2018neural,li2020scalable,tzen2019neural}. More recently, physics inspired approaches have been proposed to design algorithms for symbolic regression purposes that exploit the underlying symmetries in the problem. For example,  \cite{udrescu2020ai} show that neural networks can be used to learn reduce to the search space by quickly finding such symmetries in the dataset. Similar ideas were also applied in the context of extending the range of generators for GAN/VAE with strategies that explicitly prevent mode collapse. Wasserstein based loss functions are often preferred if the DGM suffers from mode collapse during training \cite{arjovsky2017wasserstein}.  In fact, from a technical perspective, it is possible to show that Neural SDEs has the same expressive power as high dimensional GANs \cite{kidger2021neural}. The equivalence is achieved by simply parameterizing the (drift and diffusion functions of) forward process \eqref{eqn:forward_sde} and its reverse process using separate neural networks. This connection is mathematically interesting because it allows using such DGMs for solving classical statistical problems such as Monto Carlo simulations \cite{van2021monte}.

From a fixed generator, iterative refinement techniques were proposed for dense tasks, especially at high resolution \cite{saharia2021image}.  Score based models generalize this idea, and achieve the state of the art results from many image synthesis problems \cite{dhariwal2021diffusion}. An interesting idea for Probabilistic Time Series Imputation was proposed by \cite{tashiro2021csdi}. Authors explicitly train for imputation and can exploit correlations between observed values, unlike general score-based approaches. To extend the class of distributions that can be modeled within such as framework various categorical and discrete parameterizations of the distributions have been proposed. For example, \cite{austin2021structured} introduced Discrete Denoising Diffusion Probabilistic Models, which is based on  corruption with transition matrices that mimic Gaussian kernels in continuous space, but based on nearest neighbors in embedding space, by utilizing absorbing states of a process.

On the VAE side, to deal with additional computational overhead of solving differential equations within score based models, authors in \cite{gorbach2017scalable} propose a scalable variational inference framework. Using coordinate descent on existing gradient matching approach, they propose new gradient matching algorithm that  infers states and parameters in an alternating fashion, thus offering computational speedups in certain regimes. This is somewhat closer to our work than others since we infer the parameters of $E_{\tau}$ and $D_{\tau}$ during training which correspond to state dependent parameters in their setting. By relating ODEs with Gaussian Processes \cite{wenk2019fast} provide a fast gradient matching procedure.

\section{Discussion about Improving range of Models using SDEs}
\label{sec:discussion_sde}

\paragraph{SDEs for Modeling Diversity.} In order to improve the range of generators as desired, strategies that involve explicit modeling of the {\em forward} or the corruption process have been suggested. Mathematically speaking, this can be done by using Stochastic Differential Equations (SDE) of the form shown in \cref{eqn:forward_sde}.

For generation, $\x(0)$ represents the (clean) dataset to be generated, given samples $\x(T)$ for some sufficiently large $T$. We know that under standard assumptions on $f$, we have that: as $T\uparrow \infty$, $\x(T)$ mostly represents noise, that is, has least information, . Correspondingly, given samples $\x(0)$, the learning problem is to train a generator $G$ so that it produces a randomized version of $x(0)$ from random noise $\x(\infty)$. The key property of SDEs that is attractive for distribution $G$ is that, each equilibrium solution of SDE in \eqref{eqn:forward_sde} will have its own {\em unique} trajectory, which can be used during training. We do so by coupling the intermediate iterate $x_{\tau}$ during training with specific layers in a DGM, as shown in \cref{fig:flow_arch}.

\paragraph{Training Efficiency Motivations.} When $x(T)$ defined in equation \eqref{eqn:forward_sde} can be evaluated in closed-form, then such corruption process can readily be used within the so-called {\em score} based DGM, see Section 3.3 in \cite{song2021scorebased}. We consider a challenging problem setting in this paper where we are explicitly given a non-affine corruption process via {\em discretization} which needs to be reversed.

\section{Euler–Maruyama Algorithm to aid Data Augmentation}\label{sec:applications}
Here we show the application of EM Algorithm on two simulatable processes, Lotka-Volterra \cite{kelly2016rough} and Arnold's Cat Map \cite{bao2012period}. We use 
scipy.integrate.solve\_ivp \cite{2020SciPy-NMeth} to integrate our differential equations and get the trajectory for each pixel in the image and thereby finding the trajectory of the image as a whole. The resulting images are then plotted using Matplotlib \cite{Hunter:2007}. An image of a Cat following the trajectory of both the differential equations can be seen in Figure \ref{fig:augment}.

\section{Experiments}\label{sec:expssupp}
In this section we showcase additional results from the experiments performed to explore the applications of IGO.

\subsection{Generative PCA with IGO}

\subsubsection{Implementation Details} 
\begin{figure*}[!h]
    \centering
    {\includegraphics[height = 15cm]{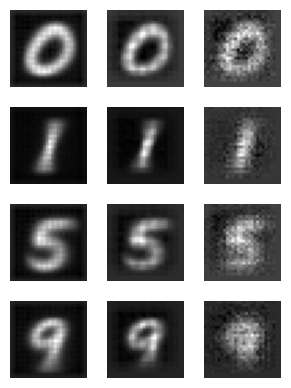} }
\caption{Reconstructed Images for Generative PCA with IGO.  \textbf{First row:} from $Z_{T}$,  \textbf{Second row:} from $Z_{\tau}$,  \textbf{Third row:} from $Z_{\tau\prime}$  }
       \label{fig:mnist_recon}%
\end{figure*}

\begin{figure*}[!h]
    \centering
    \subfloat{\includegraphics[width=0.45\linewidth]{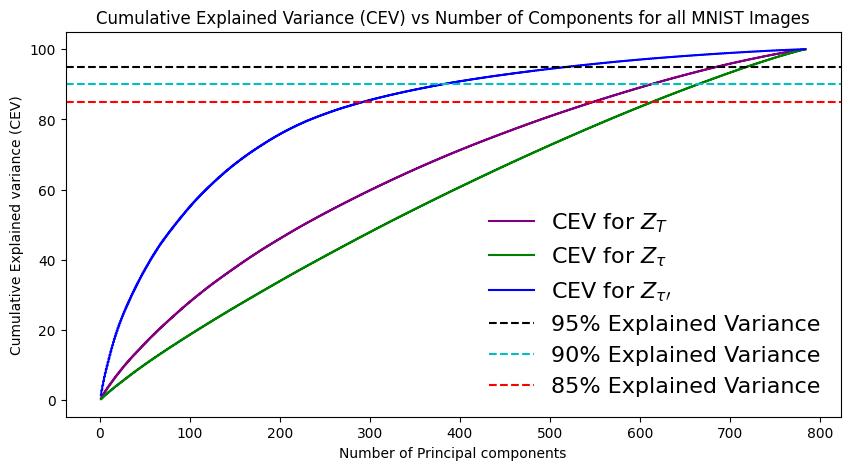} }
    \subfloat{\includegraphics[width=0.45\linewidth]{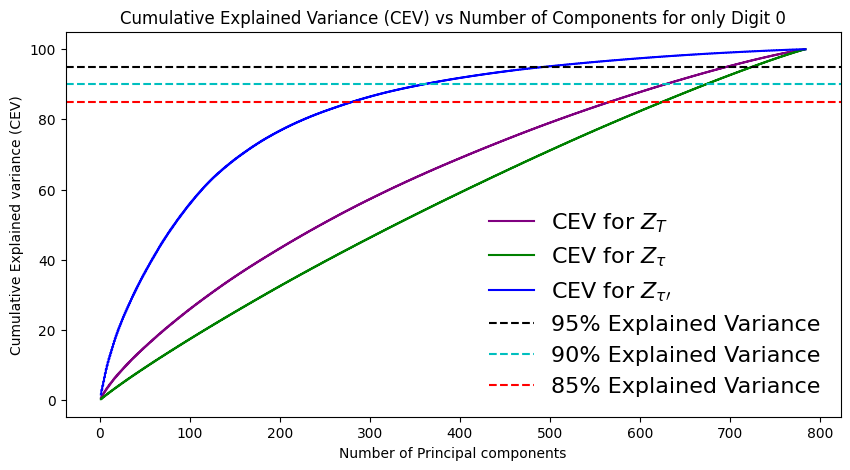} }%
    \caption{Number of Principal Components required for Explaining Variance. \textbf{Left:} For all MNIST Digits, \textbf{Right:} Only for Digit 0.}%
    \label{fig:variance}%
\end{figure*}

We show the reconstructed images for digits 0, 1, 5 and 9 from $Z_{T}$, $Z_{\tau}$ and $Z_{\tau\prime}$ in \cref{fig:mnist_recon}. \cref{fig:variance} shows the graphs comparing the explanation of variance against the Number of Principal Components. We see similar trends in the Cumulative Explained Variance, while using all the MNIST digits used together as well as while using only Digit 0.

We also provide a python script which implements the projection step to optimize the noise vectors in all the 3 different cases mentioned in our main paper ($Z_{T}$, $Z_{\tau}$ and  $Z_{\tau\prime}$). We used a generative model trained on MNIST images to perform the projection step, but the script can be utilized to do the same for any generative model.

\subsection {IGO for training Diffusion models}

\subsubsection{Generated Samples for CelebA dataset}
\begin{figure*}[!t]
    \centering
    {\includegraphics[width=\columnwidth, height=14 cm]{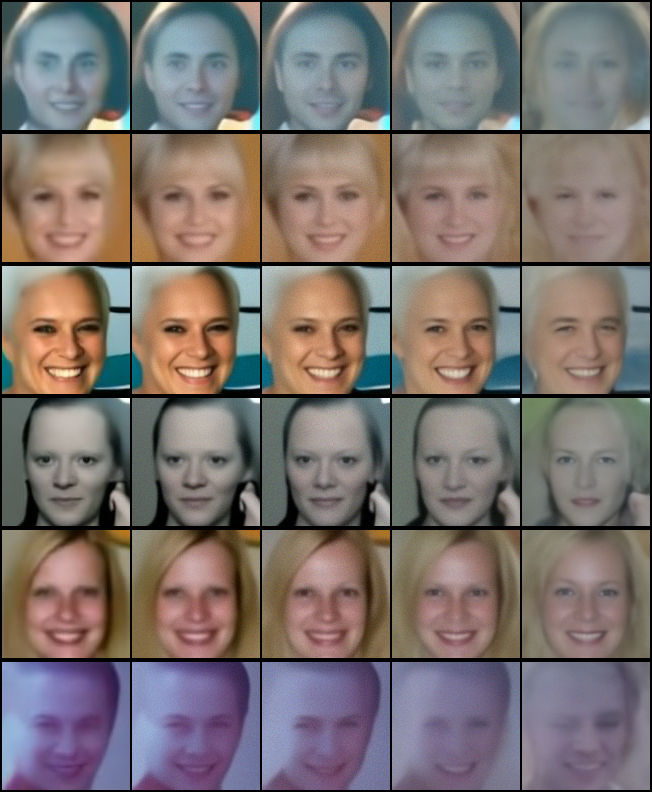} }
\caption{Samples generated from the Intermediate Layer (the rightmost image being the closest to noise.)}%
       \label{fig:inter_celeba}%
\end{figure*}

\begin{figure*}[!t]
    \centering
    {\includegraphics[width=\columnwidth, height=14 cm]{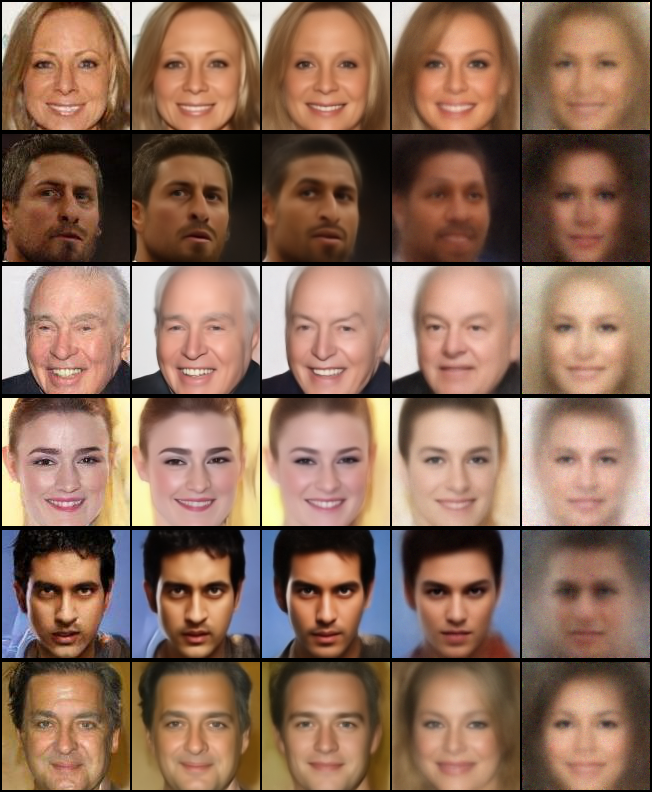} }
\caption{Samples generated from the Final Layer (the rightmost image being the closest to noise.)}%
       \label{fig:final_celeba}%
\end{figure*}

The generated samples using the progressive generation used by \cite{ho2020denoising} can be seen in \cref{fig:inter_celeba} and 
\cref{fig:final_celeba}. It can be observed from the samples in \cref{fig:inter_celeba} that we utilize our intermediate layer route to sample images from a lesser starting corruption point than the final layer.

\begin{figure*}[!h]
\centering
\begin{lstlisting}[language=Python,
        showstringspaces=false,
        formfeed=newpage,
        tabsize=4,
        commentstyle=\itshape,
        basicstyle=\footnotesize\ttfamily,
        morekeywords={sampler_normal, torch},
        frame=lines,
        title={Encoder structure for IGO on MNIST},
        label={code:mc_direct}
        ]
encoder = nn.Sequential(
        nn.Conv2d(1, 32, ks = (3,3),
                  stride=(1,1)),
        nn.Conv2d(32, 64, ks = (3,3),
                  stride=(2,2)),
        
        #Intermediate Encoder   
        nn.Conv2d(1, 64, ks = (5,5),
         stride=(2,2)),
         
        nn.Conv2d(64, 128, ks = (3,3),
                  stride=(2,2)),
                  
        nn.Conv2d(128, 256, ks = (3,3),
                  stride=(2,2))
        )
\end{lstlisting}
\begin{lstlisting}[language=Python,
        showstringspaces=false,
        formfeed=newpage,
        tabsize=4,
        commentstyle=\itshape,
        basicstyle=\footnotesize\ttfamily,
        morekeywords={sampler_normal, torch},
        frame=lines,
        title={Decoder structure used for IGO on MNIST},
        label={code:mc_direct}
        ]

decoder = nn.Sequential(
        nn.ConvTranspose2d(256, 128, ks = (3,3),
                  stride=(2,2)),
        nn.ConvTranspose2d(128, 64, ks = (3,3),
                  stride=(2,2)),
        
        #Intermediate Decoder
        nn.ConvTranspose2d(64, 1, ks = (6,6),
                  stride=(2,2)),
        
        nn.ConvTranspose2d(64, 32, ks = (3,3),
                  stride=(2,2)),
        
        nn.ConvTranspose2d(32, 1, ks = (3,3),
                  stride=(1,1))
        )
\end{lstlisting}
\caption{Implementing IGO on Pytorch involves encoding intermediate $\tx$ with $E_{\tau}$, which eventually gets decoded by $D_{\tau}$.}
\label{fig:impl}
\end{figure*}

\subsubsection{Implementation Details for MNIST dataset} 
\label{sec:mnistsupp}
To analyze the effect of $\alpha$, we perform another experiment similar to \cref{sec:igo-unet}, using MNIST dataset. During the forward pass, our simulated iterates $\x_{t}$ and $\tx_{\tau}$ ({$\tau = t/2$}) are passed into the network, as shown in \cref{fig:flow_arch}. We use a convolution layer with a kernel of size 5 and stride 2 as the intermediate encoder $E_{\tau}$ and a de-convolution layer with
a kernel of size 6 and stride 2 as the intermediate decoder
$D_{\tau}$ . The training is done using a convex combination of
loss, same as done in the case of CelebA dataset. Figure \ref{fig:impl} summarises the PyTorch implementation of IGO in a U-net based Generative setting for the MNIST dataset.

\subsubsection{Generated Samples for MNIST dataset} 
\begin{figure*}[!h]
\centering
\includegraphics[width=0.4\textwidth]{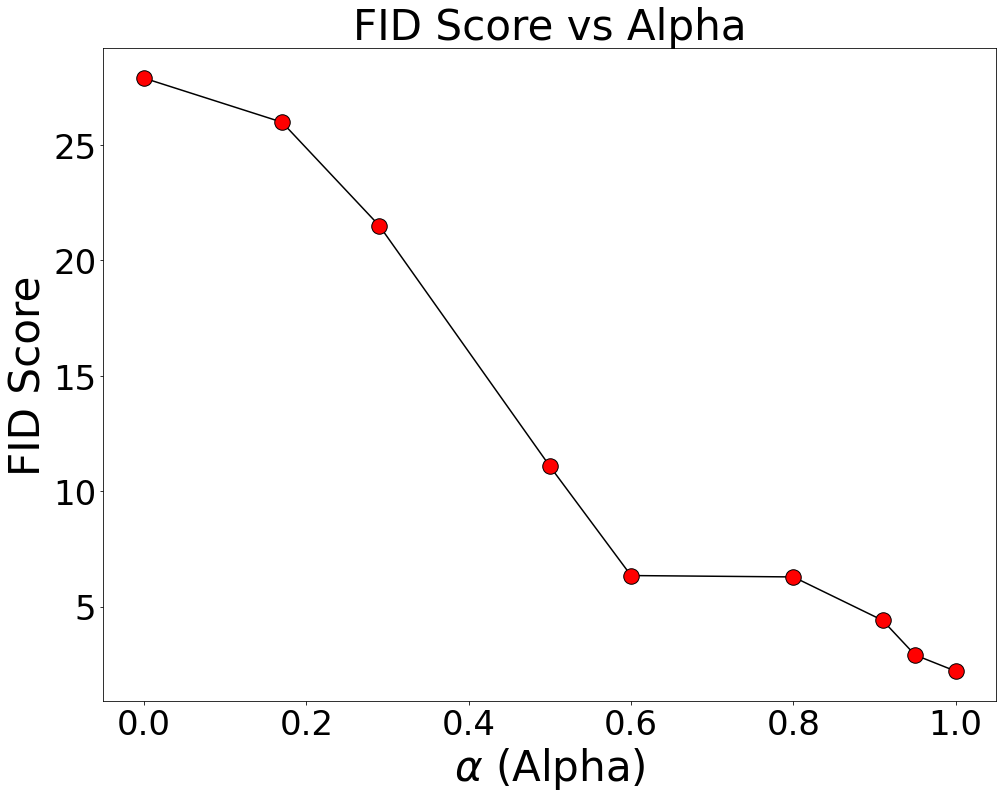}
\caption{FID scores for Images generated using models trained using different values of $\alpha$
\label{fig:fid}}
\end{figure*}

 For quantitative comparisons of the images generated from models using different values of $\alpha$, we utilized the pytorch implementation of Fretchet Inception Distance (FID) \cite{NIPS2017_8a1d6947} score provided by \cite{Seitzer2020FID}. The results can be seen in \cref{fig:fid}. We can see that the values of $\alpha$ can be chosen depending on the trade off between range expansion and the FID. 
 
\begin{figure*}[]
    \centering
    {\includegraphics[width=\columnwidth]{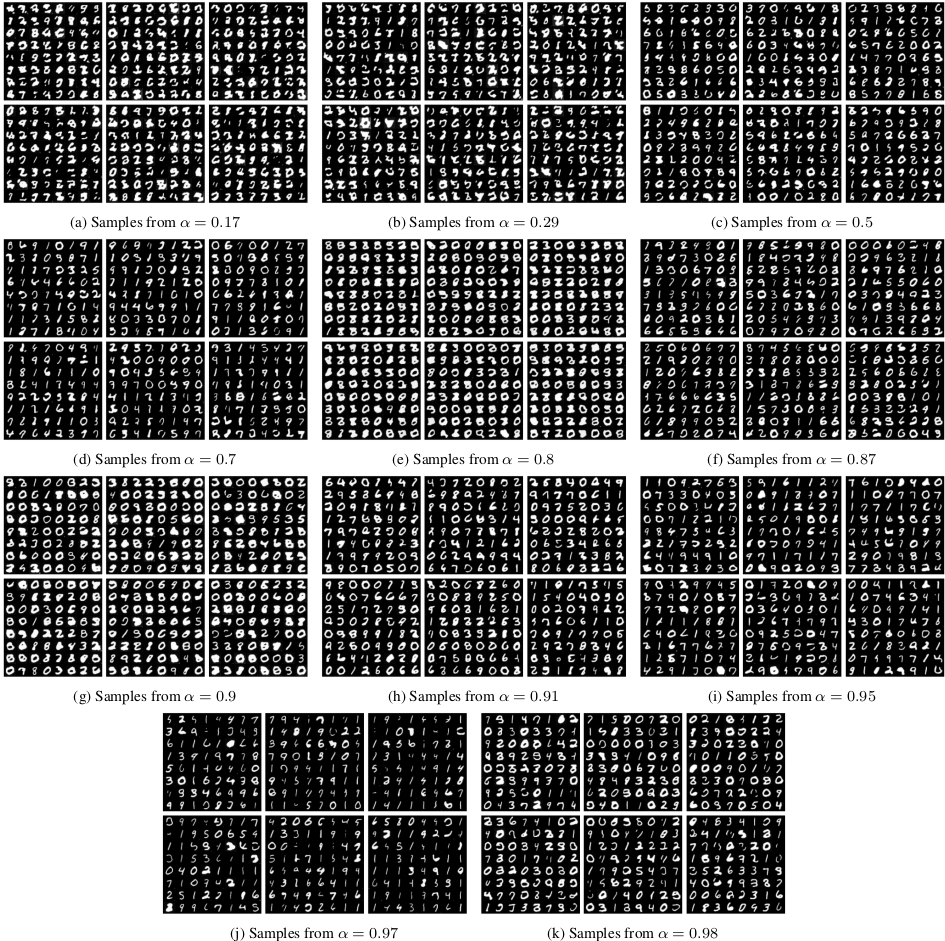} }
\caption{Samples for different values of $\alpha$}%
       \label{fig:mnistimages}%
\end{figure*}

 \cref{fig:mnistimages} shows the Generated samples with the perturbation kernel using Arnold's Cat Map as its drift function, for different values of $\alpha$. The samples generated do have some structural deformities, indicating that the range of the generators is improved due to our setup. Moreover, note that our IGO scheme is robust with respect
to $\alpha$ as shown by the gradual degradation in performance as we put more weight on the gradients provided by the intermediate iterates. Thus suggesting that using a combinatorial objective function can help increase the range of a generative model.

\subsection{IGO on Panel Data}
 \label{sec:panelsupp}
\subsubsection{Implementation Details} 
 Experiments were run on NVIDIA 2080ti GPU for 300 epochs. First 100 epochs were used to train encoder/decoder only, without any temporal component. 
For Rotating MNIST (2d data) the encoder-decoder model that we used in our experiments are described in Figure \ref{fig:2d_encoder}.

\subsection{IGO in 3D Processing}
\label{sec:igo-3dsupp}
\subsubsection {Qualitative Results}
\begin{figure*}[!h]
    \centering
    \subfloat{\includegraphics[width=5.5 cm]{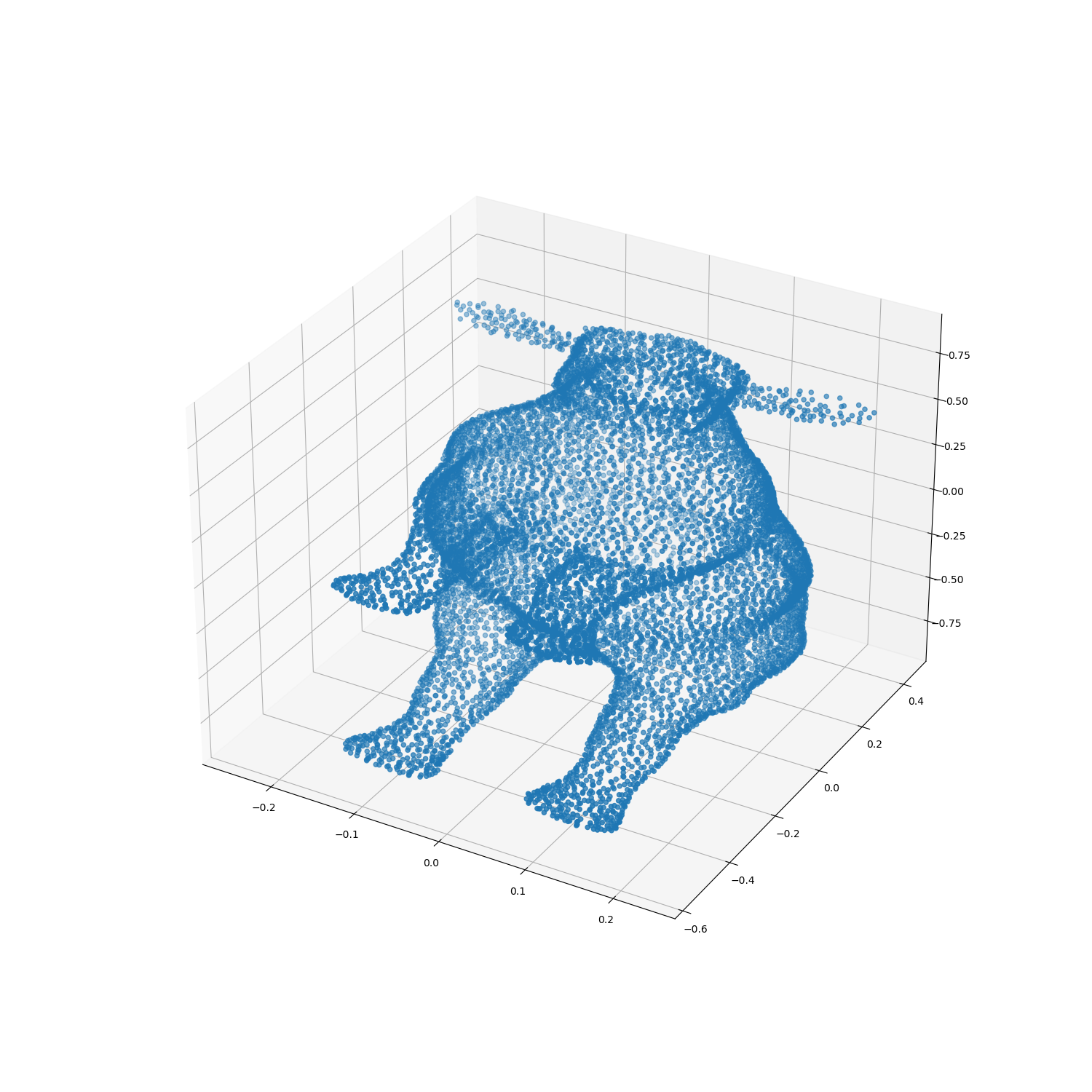} }
    \subfloat{\includegraphics[width=5.5 cm]{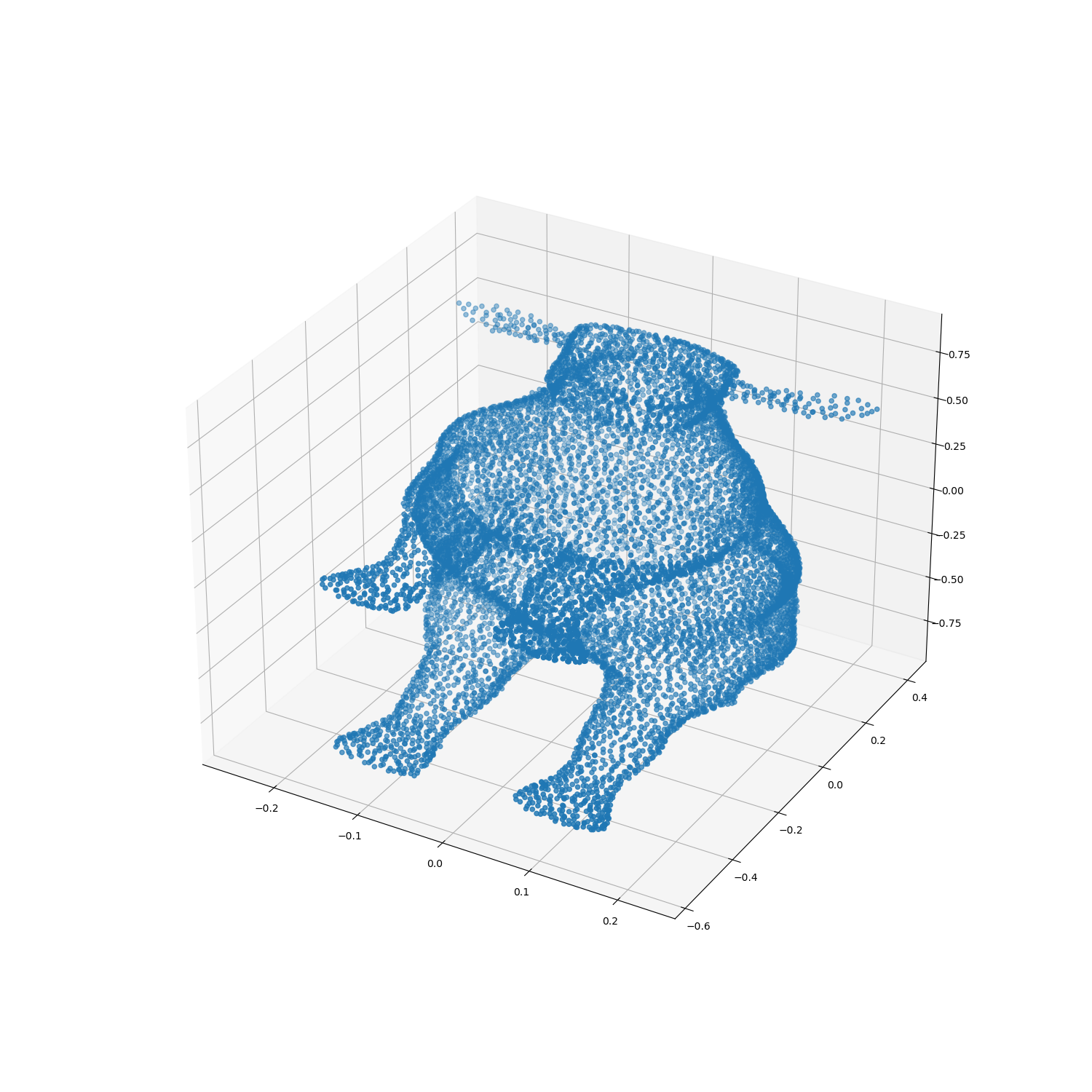} }%
    \caption{Denoised Image using a) Intermediate layer b) Final layer, $\alpha = 0.98$}%
    \label{fig:example3}%
\end{figure*}

\begin{figure*}[!h]
    \centering
    \subfloat{\includegraphics[width=5.5 cm]{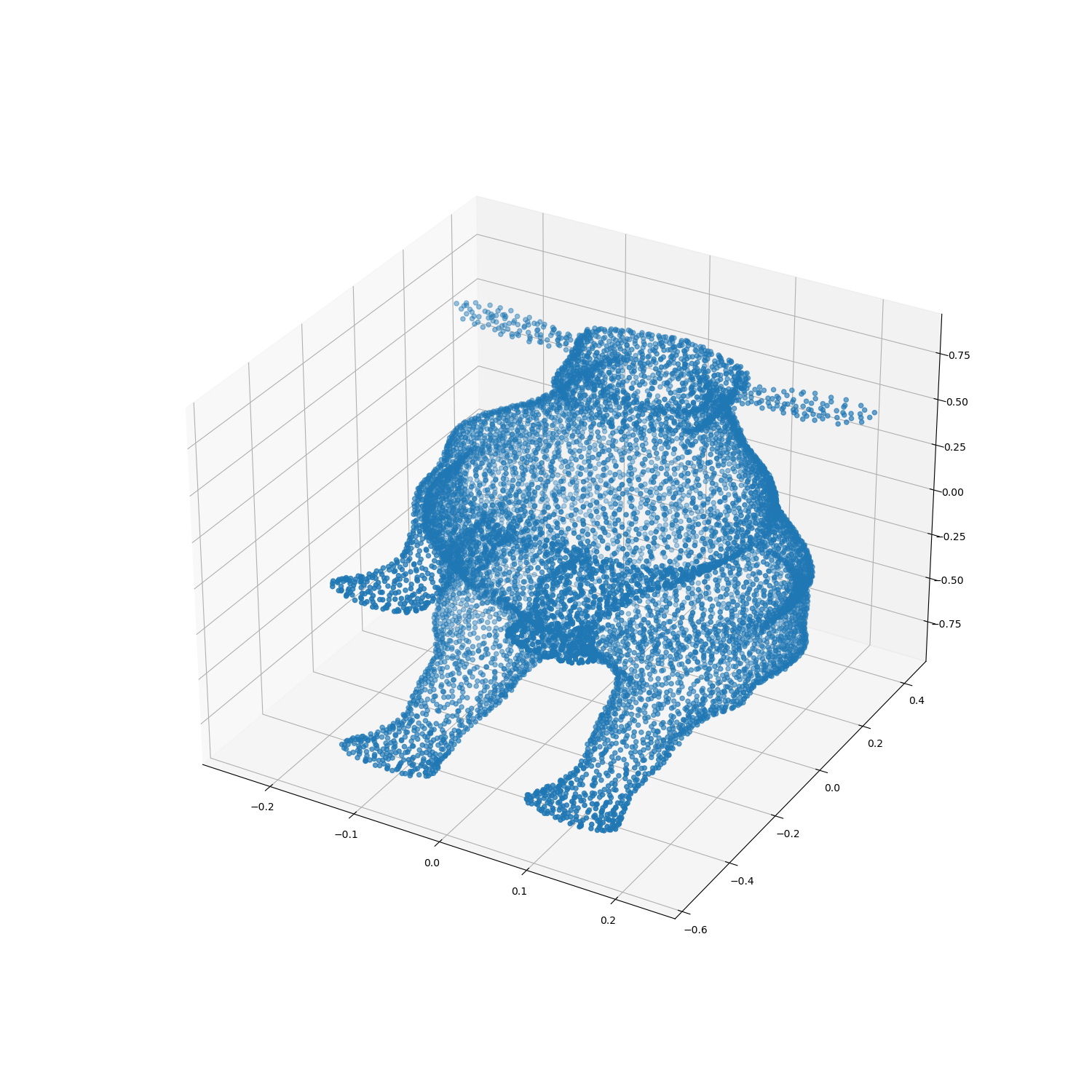} }
    \subfloat{\includegraphics[width=5.5 cm]{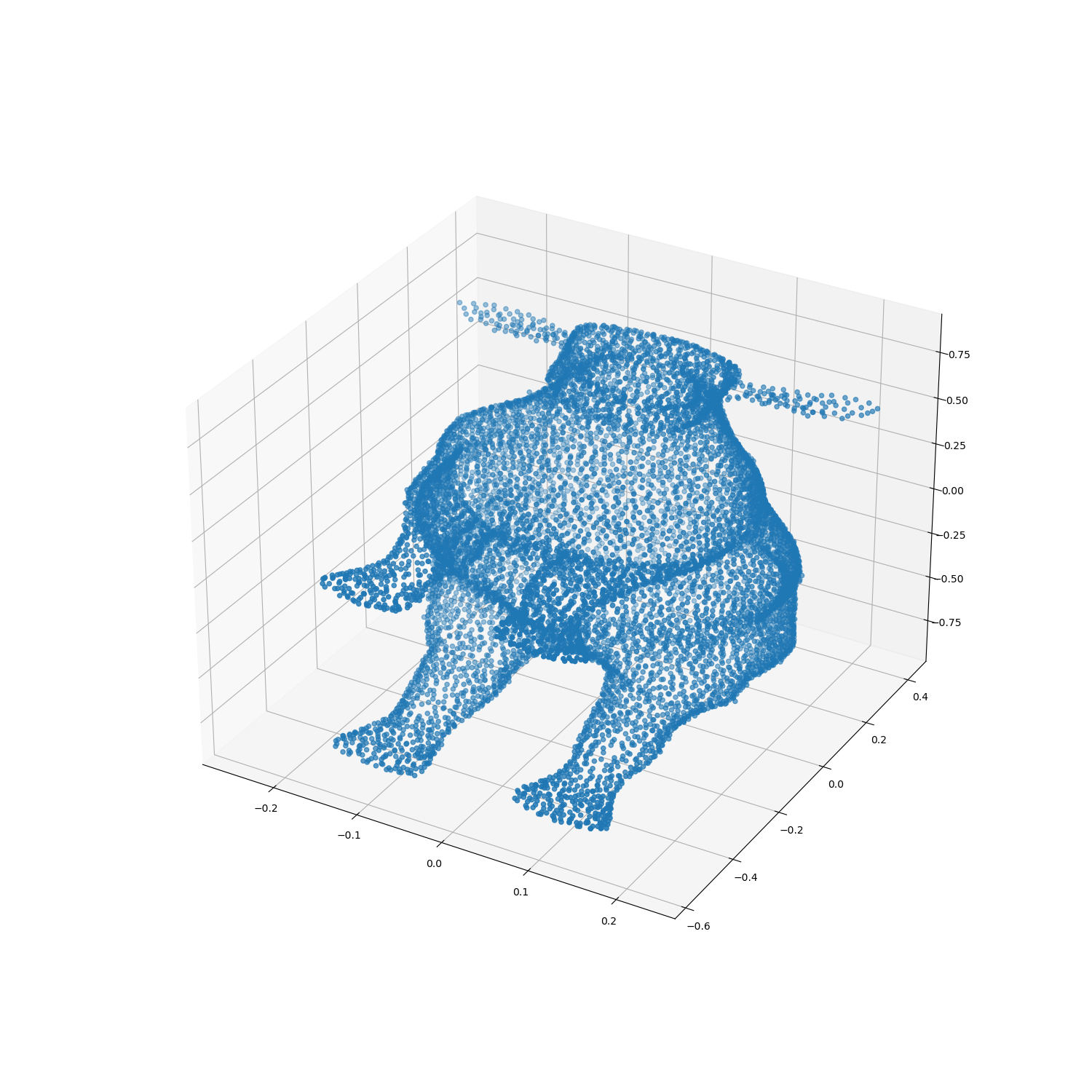} }%
    \caption{Denoised Image using a) Intermediate layer b) Final layer, $\alpha = 0.9$}%
    \label{fig:example4}%
\end{figure*}

\begin{figure*}[!h]
    \centering
    \subfloat{\includegraphics[width=5.5 cm]{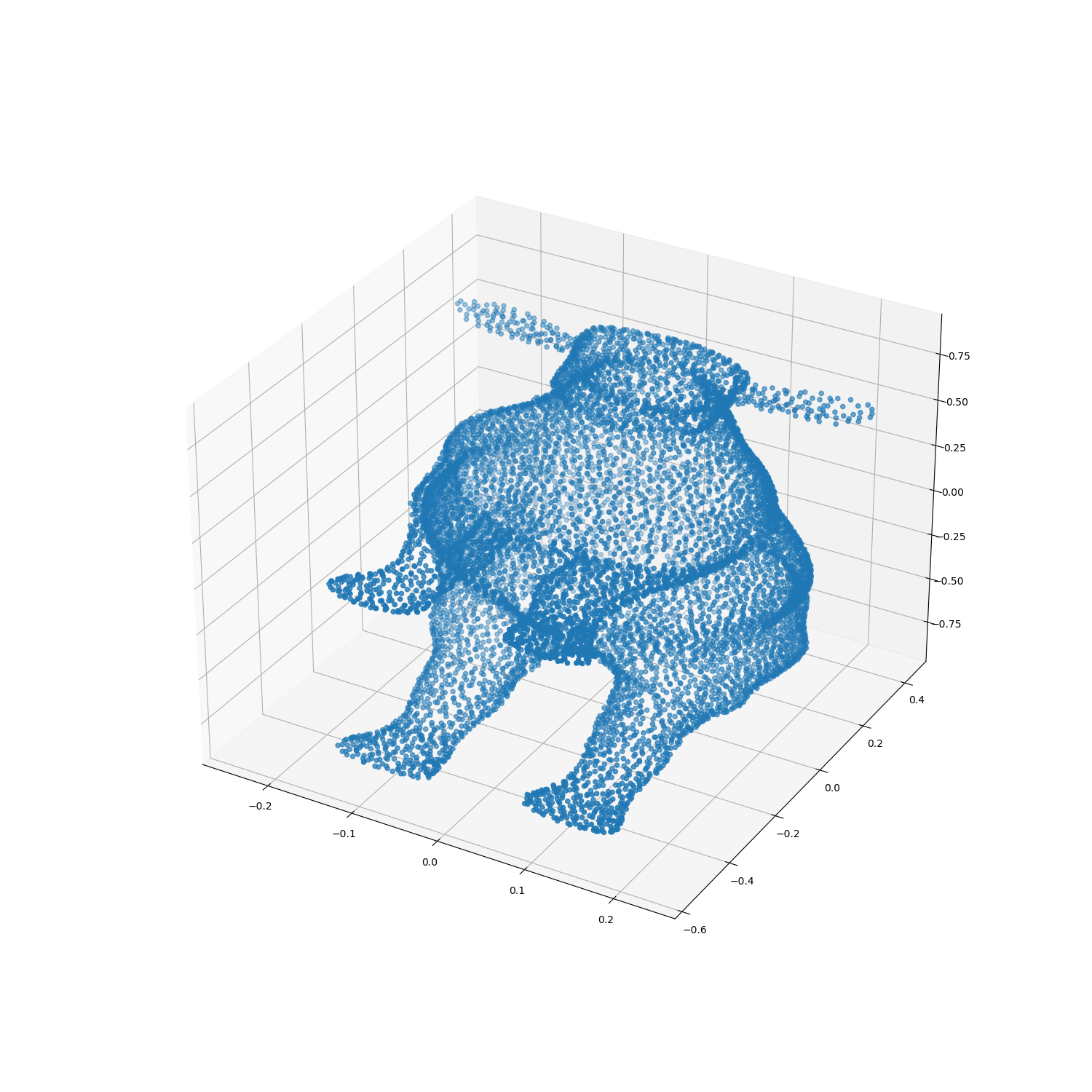} }
    \subfloat{\includegraphics[width=5.5 cm]{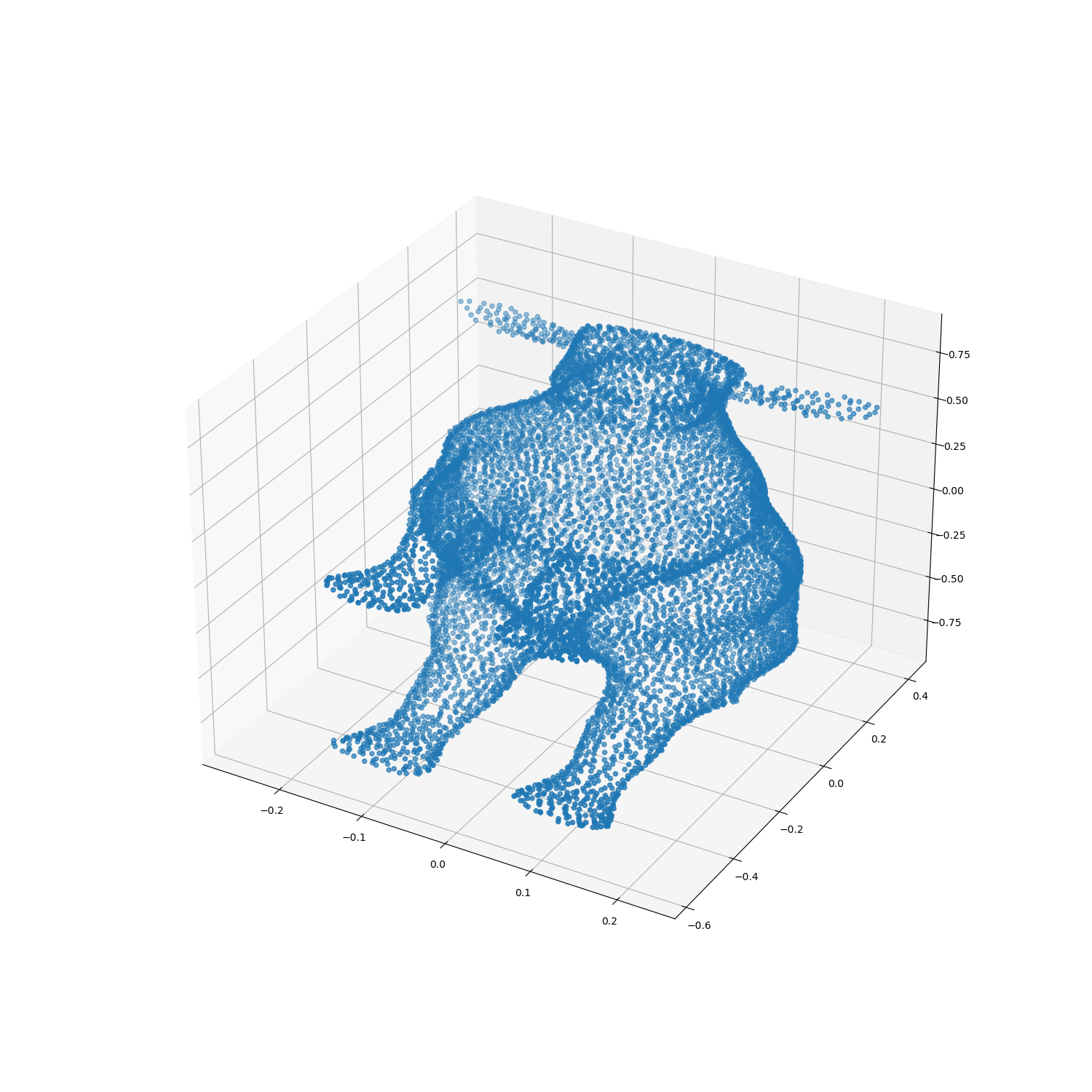} }%
    \caption{Denoised Image using a) Intermediate layer b) Final layer, $\alpha = 0.5$}%
    \label{fig:example5}%
\end{figure*}

Figure \ref{fig:example3}, \ref{fig:example4} and \ref{fig:example5} show the qualitative results from the experiment in \cref{sec:pcd-exps}, for different values of the hyperparameter $\alpha$.

\begin{figure*}[!h]
\centering
\begin{lstlisting}[language=Python,
        showstringspaces=false,
        formfeed=newpage,
        tabsize=4,
        commentstyle=\itshape,
        basicstyle=\footnotesize\ttfamily,
        morekeywords={sampler_normal, torch},
        frame=lines,
        title={Encoder structure used in ROTATING MNIST},
        label={code:mc_direct}
        ]
encoder = nn.Sequential(
        nn.Conv2d(input_dim, 12, ks,
                  stride=1, padding=1),
        nn.ReLU(),
        nn.Conv2d(12, 24, ks,
                  stride=2, padding=1),
        nn.ReLU(),
        nn.Conv2d(24, output_dim, ks,
                  stride=2, padding=1),
        nn.Flatten(2),
        nn.Linear(49, 1),
        nn.Flatten(1)
        )
\end{lstlisting}
\begin{lstlisting}[language=Python,
        showstringspaces=false,
        formfeed=newpage,
        tabsize=4,
        commentstyle=\itshape,
        basicstyle=\footnotesize\ttfamily,
        morekeywords={sampler_normal, torch},
        frame=lines,
        title={Decoder structure used in ROTATING MNIST},
        label={code:mc_direct}
        ]
extend_to_2d = nn.Linear(input_dim,
                         49 * input_dim)
decoder = nn.Sequential(
        nn.ConvTranspose2d(input_dim,
                           24,
                           ks,
                           stride=2,
                           padding=1,
                           output_padding=1),
        nn.ConvTranspose2d(24,
                           12,
                           ks,
                           stride=2,
                           padding=1,
                           output_padding=1),
        nn.ConvTranspose2d(12, output_dim, ks,
                           stride=1, padding=1),
        nn.Sigmoid(),
        )
\end{lstlisting}
\caption{Description of Encoder and Decoder used in experiment with 2d data structure: Rotating MNIST.}
\label{fig:2d_encoder}
\end{figure*}

\begin{table*}[!h]
\centering
{%
\begin{tabular}{l|cc|cc|cc|cc}
\specialrule{1pt}{1pt}{0pt}
\multicolumn{1}{c|}{\# Points} & \multicolumn{4}{c|}{10K (Sparse)} & \multicolumn{4}{c}{50K (Dense)} \\
\hline
\multicolumn{1}{c|}{Noise} & 
\multicolumn{2}{c|}{1\%} & \multicolumn{2}{c|}{2\%} &
\multicolumn{2}{c|}{1\%} & \multicolumn{2}{c|}{2\%} \\
\hline
     \multicolumn{1}{c|}{Model} &
     CD & P2M & CD & P2M &
     CD & P2M & CD & P2M \\
\specialrule{1pt}{0pt}{0pt}
     Bilateral \cite{10b55a273a81438ca33320cac02aafb8} &
        3.646 & 1.342 & 5.007 & 2.018 &
        0.877 & 0.234 & 2.376 & 1.389 
        \\
    Jet \cite{cazals:inria-00097582}   &
        2.712 & 0.613 & 4.155 & 1.347 &
        0.851 & 0.207 & 2.432 & 1.403
        \\
     MRPCA \cite{Mattei2017PointCD} &
        2.972 & 0.922 & 3.728 & 1.117 &
      \bf  0.669 & \bf 0.099 & 2.008 & 1.033
        \\
     GLR \cite{GLR}   &
        2.959 & 1.052 & 3.773 & 1.306 &
        0.696 & 0.161 & 1.587 & 0.830 
        \\

     PCNet \cite{rakotosaona2020pointcleannet}   &
        3.515 & 1.148 & 7.467 & 3.965 &
        1.049 & 0.346 & 1.447 & 0.608
        \\
     DMR \cite{luo2020DMR}  &
        4.482 & 1.722 & 4.982 & 2.115 &
        1.162 & 0.469 & 1.566 & 0.800
        \\
     Score-Based PCD \cite{Luo_2021_ICCV} &
    \bf  2.521 & \bf 0.463 & \bf 3.686 & \bf  1.074 &
    0.716 & 0.150 & \bf 1.288 & \bf 0.566
    \\
 \cmidrule{1-9}
     Ours ($\alpha = 0.8$) &
         2.710 &  0.593 &  4.292 &  1.524 &
         0.954 &  0.320 &  2.456 &  1.517 
        \\
\bottomrule
\end{tabular}
}
\caption{\label{tab:table_data}Comparison against other denoising algorithms. The CD as well as the P2M scores are multiplied by $1e{+4}$.}
 \vspace{-15pt}
\end{table*}

\subsubsection {Quantitative Results}
The comparison with the baselines for a noise levels $1\%$ and $2\%$ are shown in Table \ref{tab:table_data}, for the model trained with $\alpha = 0.8$. We can see that our model's performance is close to the state of the art model, in the lower noise settings, while beating a lot of other baselines.

\end{document}